\documentclass[]{fairmeta}

\usepackage{comment}
\usepackage[inkscapelatex=false]{svg}
\usepackage{multirow}
\usepackage{amssymb}
\usepackage{pifont}
\usepackage{colortbl}
\usepackage{caption}
\usepackage{enumitem}
\usepackage{booktabs}
\usepackage{amsmath}
\usepackage{appendix}
\usepackage{siunitx}
\usepackage{tabularray}
\usepackage{pgf}
\usepackage{makecell}
\usepackage{siunitx}

\newcommand{\ignore}[1]{}
\definecolor{greenish}{RGB}{217,234,211}
\colorlet{greenlight}{greenish!30!white}
\definecolor{cyanish}{RGB}{66,245,239}
\colorlet{cyanlight}{cyanish!30!white}
\definecolor{blueish}{RGB}{66,135,245}
\colorlet{bluelight}{blueish!30!white}
\definecolor{turquoisish}{RGB}{208,224,227}
\colorlet{turquoiselight}{turquoisish!30!white}
\definecolor{redish}{RGB}{244,205,204}
\colorlet{redlight}{redish!30!white}
\definecolor{orangish}{RGB}{252,228,205}
\colorlet{orangelight}{orangish!30!white}
\definecolor{purplish}{RGB}{217,210,232}
\colorlet{purplelight}{purplish!30!white}
\definecolor{pinkish}{RGB}{234,209,220}
\colorlet{pinklight}{pinkish!30!white}

\newcommand{\System}{LayerSkip}
\title{\System: Enabling Early Exit Inference and Self-Speculative Decoding}

\author[1,\dagger,*]{Mostafa Elhoushi}
\author[1,\dagger,*]{Akshat Shrivastava}
\author[2,\dagger]{Diana Liskovich}
\author[1]{Basil Hosmer}
\author[2]{Bram Wasti}
\author[3]{Liangzhen Lai}
\author[4]{Anas Mahmoud}
\author[1]{Bilge Acun}
\author[6]{Saurabh Agrawal}
\author[7]{Ahmed Roman}
\author[3]{Ahmed A Aly}
\author[1,5]{Beidi Chen}
\author[1]{Carole Jean-Wu}

\affiliation[1]{FAIR at Meta}
\affiliation[2]{GenAI at Meta}
\affiliation[3]{Reality Labs at Meta}
\affiliation[4]{University of Toronto}
\affiliation[5]{Carnegie Mellon University}
\affiliation[6]{University of Wisconsin-Madison}
\affiliation[7]{Dana-Farber Cancer Institute}

\contribution[*]{Equal Contribution}
\contribution[\dagger]{Core Contributor}

\abstract{We present \System, an end-to-end solution to speed-up \ignore{training (both pre-training or finetuning) as well as} inference of large language models (LLMs). First, during training we apply layer dropout, with low dropout rates for earlier layers and higher dropout rates for later layers, and an early exit loss where all transformer layers share the same exit. Second, during inference, we show that this training recipe increases the accuracy of early exit at earlier layers, without adding any auxiliary layers or modules to the model. Third, we present a novel self-speculative decoding solution where we exit at early layers and verify and correct with remaining layers of the model. Our proposed self-speculative decoding approach has less memory footprint than other speculative decoding approaches and benefits from shared compute and activations of the draft and verification stages. We run experiments on different Llama model sizes on different types of training: pretraining from scratch, continual pretraining, finetuning on specific data domain, and finetuning on specific task. We implement our inference solution \ignore{ on GPU and CPU, as well as on phone} and show speedups of up to 2.16$\times$ on summarization for CNN/DM documents, 1.82$\times$ on coding, and 2.0$\times$ on TOPv2 semantic parsing task.} 

\date{\today}
\correspondence{Mostafa Elhoushi, Akshat Shrivastava \\ \null\quad\quad\quad\quad\quad\quad at \email{melhoushi@meta.com}, \email{akshats@meta.com}}

\metadata[Code] {\url{https://github.com/facebookresearch/LayerSkip}}

\begin{document}

\maketitle

\section{Introduction}
\label{sec:intro}

Large Language Models (LLMs) have been deployed to many applications, yet their high compute and memory requirements lead to high financial and energy costs when deployed to GPU servers~\citet{benchmarkLLM}. Acceleration solutions do exist to deploy to commodity GPUs on laptops but they suffer from significant drop in accuracy~\citet{zhu2023survey}. Accelerating LLMs further to mobile or edge devices is still an active research area~\citet{coplu2023performance,liu2024mobilellm}. While a large portion of LLM acceleration approaches reduce number of non-zero weights~\citet{xia2023flashllm} (a.k.a. sparsity), number of bits per weight~\citet{xiao2023smoothquant} (a.k.a. quantization), number of heads per layer~\citet{headpruning_layerwise} (a.k.a. head pruning), a smaller portion of approaches focus on reducing number of layers~\citet{LayerDrop,DepthAdaptive}. In this paper, we explore reducing the number of layers required for each token by exiting early during inference. Unlike quantization or sparsity, acceleration by reducing number of layers does not require specialized hardware or software kernels.

\begin{figure}[t!]
    \vskip 0.2in
    \begin{center}
    \centerline{\includegraphics[width=\columnwidth]{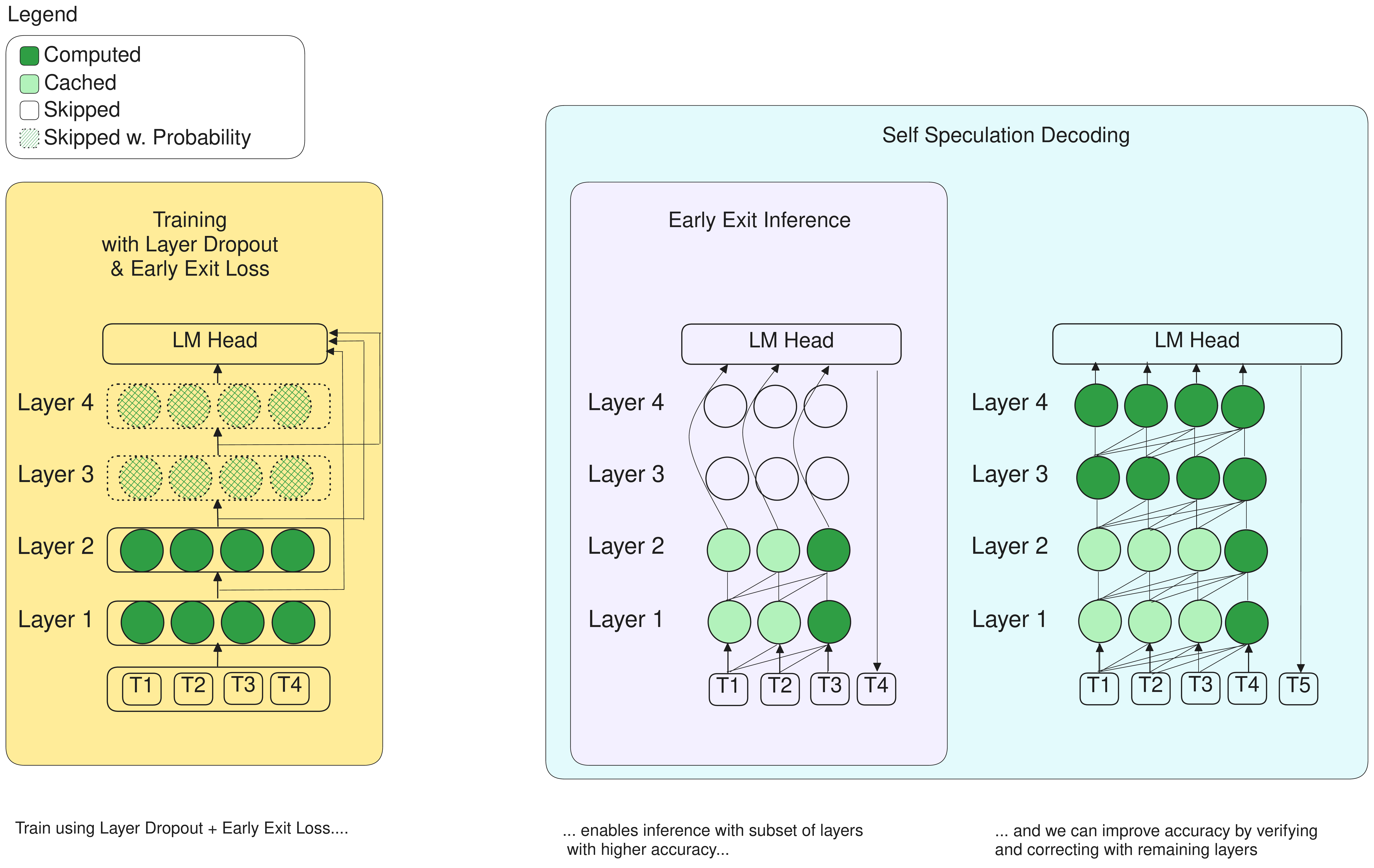}}
    \caption{Overview of our end-to-end solution, \System, showing its 3 components.}
    \label{fig:overview}
    \end{center}
    \vskip -0.2in
\end{figure}

Moreover, a popular research trend in LLM acceleration is speculative decoding~\citet{yaniv_speculative_decoding,Chen2023AcceleratingLLSpeculativeSampling} that has no drop in accuracy, where a large model, referred to as the \textit{main} model, is accompanied with a faster model, referred to as the \textit{draft} model. The advantage of speculative decoding is that it leads to faster inference compared to the main model, but requires a larger memory footprint and complexity in implementation to maintain key-value (KV) cache in two different models. In addition to exiting early, this paper also proposes combining exiting early with speculative decoding to propose a \textit{self-speculative decoding approach} that does not require an additional model or auxiliary layers.

The contribution of this paper is an end-to-end solution:
\begin{itemize}[noitemsep, nolistsep, topsep=0pt]
    \item a training recipe that combines layer dropout and early exit loss, that leads to,
    \item inference that is more robust to exiting at earlier layers of the model, essentially creating different sized sub-models within the same model, and
    \item a self-speculative decoding solution that decodes with earlier layers and verifies and corrects with later layers.
\end{itemize}

The solution achieves speedups between 1.34$\times$ and $2.16\times$ depending on the task. We provide an overview of the solution in Figure~\ref{fig:overview}.

\section{Motivation}
\label{sec:motivation}

\subsection{Exiting Earlier in LLMs}
\label{sec:motivation:early_exit}
\begin{figure*}[t]
     \centering
     \begin{subfigure}[b]{0.48\textwidth}
         \centering
         \includegraphics[width=\columnwidth]{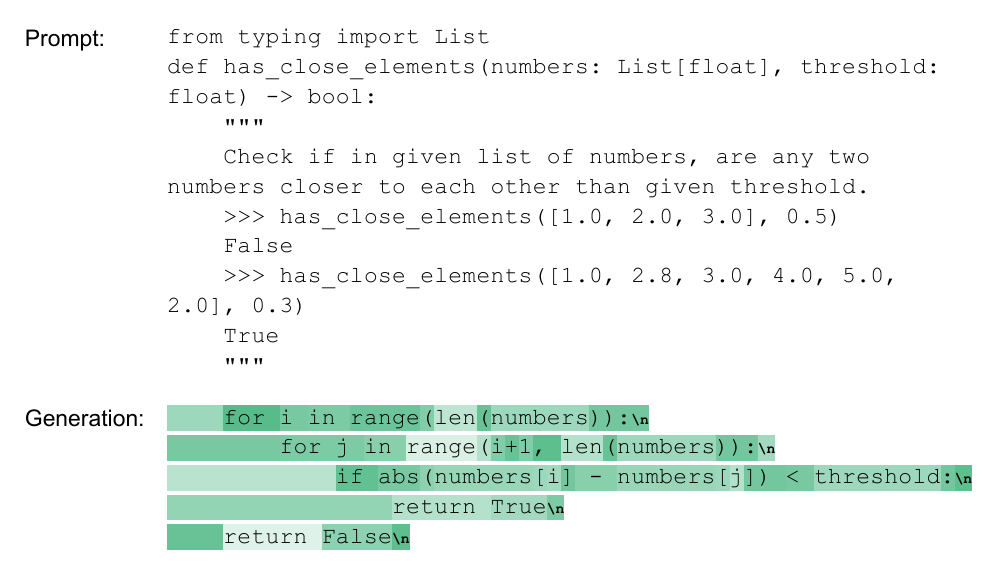}
         \caption{}
         \label{fig:llamav1-7B-generation:color-coded}
     \end{subfigure}
     \hfill
     \begin{subfigure}[b]{0.48\textwidth}
         \centering
         \includegraphics[width=\columnwidth]{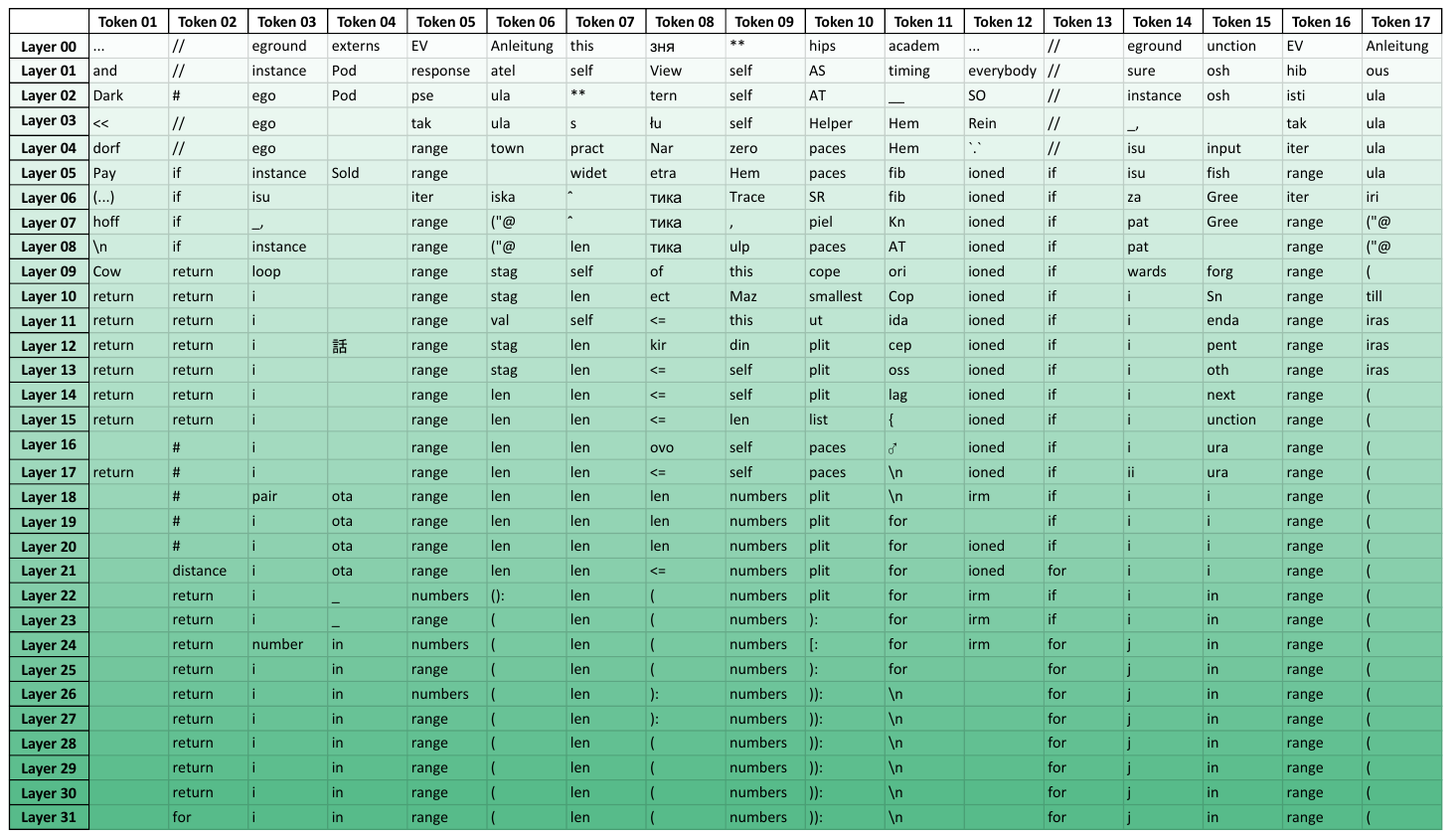}
         \caption{}
         \label{fig:llamav1-7B-generation:layerwise}
     \end{subfigure}
     \caption{(a) A prompt from the HumanEval dataset~\citet{HumanEval} and corresponding text generated by Llama1 7B. The color of each generated token corresponds to the earliest layer in the model that predicted it. (b) Token prediction at each layer in Llama1 7B.}
     \label{fig:llamav1-7B-generation}
\end{figure*}

To motivate our approach, we investigate, with an example prompt, what happens in each layer in a LLM. In Figure~\ref{fig:llamav1-7B-generation:color-coded}, we provide the first prompt from the HumanEval coding dataset~\citet{HumanEval} to a pretrained Llama1 7B model~\citet{llama1}. The prompt consists of a Python function header and a docstring, and the model autocompletes it by defining the function body. When generating each token, we probe each transformer layer in the LLM by projecting its output embeddings on the language model (LM) head (that consists of the model's final layer normalization and linear layer), applying softmax, and then obtaining the index of the output element with highest value. The resulting index corresponds to the predicted token at this layer. This operation is referred to in some literature as the unembedding operation~\citet{FormalVerificationTransformer,cancedda2024spectral}, as it converts an embedding to an index. Unembedding at each layer is equivalent to early-exit at that layer, i.e., it is equivalent to skipping the remaining transformer layers to the model's LM head.

The token predictions across layers in Figure~\ref{fig:llamav1-7B-generation:layerwise} illustrate the evolution of embeddings from an input token fed to the model to the predicted next token by the model. When analyzing the token prediction in each layer in Figure~\ref{fig:llamav1-7B-generation:layerwise}, we make a few observations. First, token predictions in earlier layers appear to be irrelevant as they correspond to the previous token projected on the model's embedding layer's weights, which are different from the weights of the LM head. In later layers, token predictions converge to the final prediction. Second, we do not always need all the layers to predict the correct token. In fact, most of the time, the final token prediction is predicted fewer layers  before the end. We also notice that intermediate layers are sometimes hesitant and ``change their minds'', e.g., for Token 05, the model was predicting ``range'' as early as Layer 07, but changed its mind between Layer 22 and Layer 26, before settling again on ``range''.

Similar analysis was done in~\citet{logitlens} and ~\citet{prediction_saturation} on a GPT2 model~\citet{gpt2} as it developed predictors to estimate when prediction saturates to exit early. For the particular example we present in Figure~\ref{fig:llamav1-7B-generation}, we find, on average, a token requires 23.45 layers out of the model's 32 layers. Hence, even if we have a perfect predictor that has zero compute overhead, we can only save up to 26\% of computation. Therefore, there is a need to make LLM models require fewer layers to predict each token, and spend less compute being hesitant or ``changing its mind''. By default, deep learning models are not motivated to predict their final output early and instead spread their compute across all layers~\citet{voita2019bottomup,voita2023neurons}. We see in Figure~\ref{fig:llamav1-7B-generation:layerwise}, that tokens we would consider easy or straightforward to predict, e.g., Token 02 that starts a for-loop, required all 32 layers to predict ``for''. We would like our model to be less reliant on later layers and only use later layers for harder tokens. We would like our models to be more reliant on earlier layers than later layers. To do that, we propose skipping layers during training, which we refer to as layer dropout. However, we use higher dropout rates for later layers and lower dropout rates for earlier layers, to make the model less reliant on later layers. 

\begin{figure*}[t]
    \vskip 0.2in
    \begin{center}
    \centerline{\includegraphics[width=0.5\columnwidth]{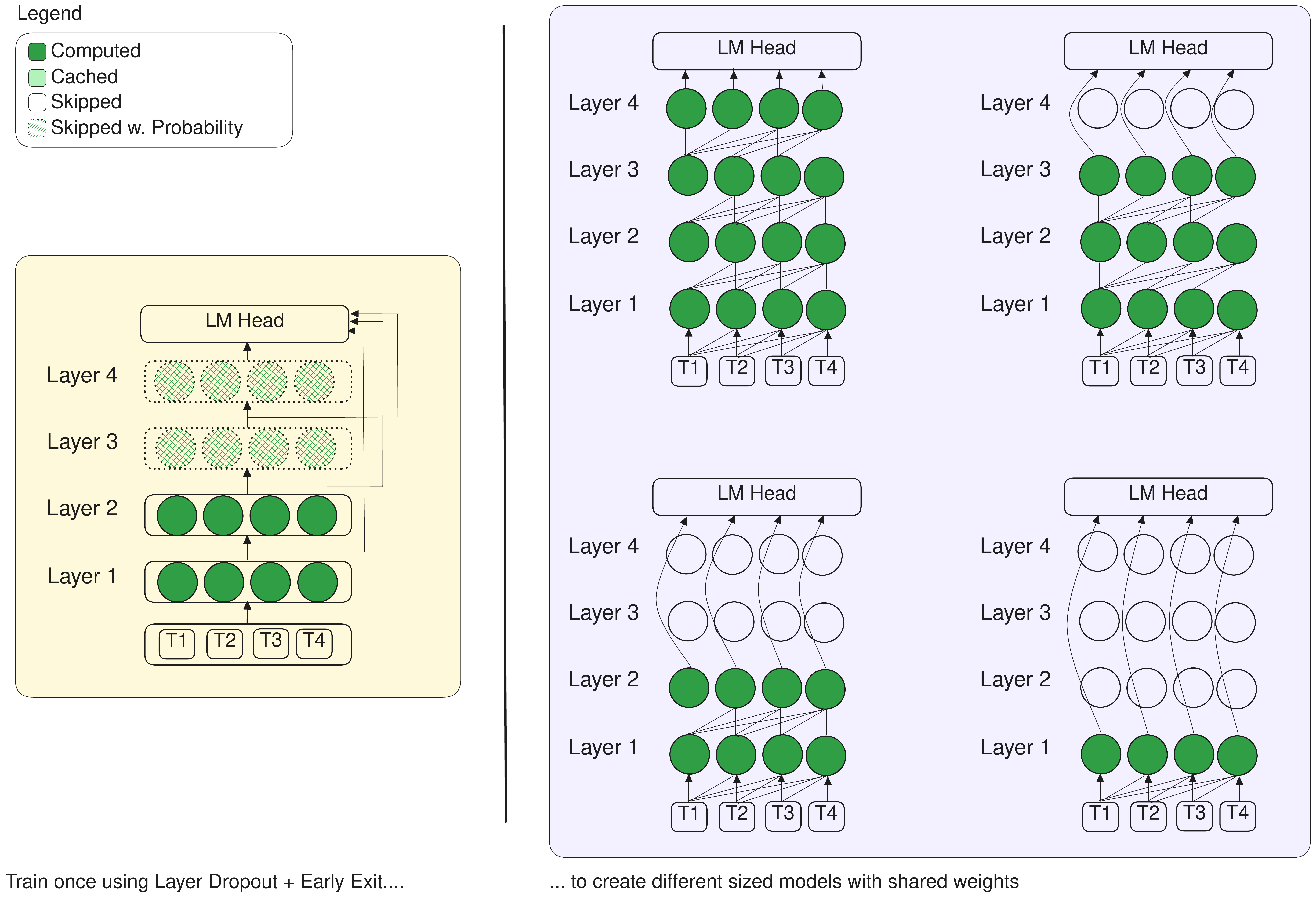}}
    \caption{We propose using layer dropout and early exit loss during training to create a model that is equivalent to an ensemble of models of various depths.}
    \label{fig:motivation}
    \end{center}
    \vskip -0.2in
\end{figure*}

Moreover, LM heads in LLMs are trained to unembed embeddings from the last transformer layer. They were not trained to unembed from earlier layers. Therefore, our solution also adds a loss function during training to make LM heads better ``understand'' embeddings of earlier layers. While most papers that explored early exit~\citet{calm,DepthAdaptive} trained a dedicated LM head for each transformer layer, and some have introduced additional modules for each early exit~\citet{SCAN}, we chose to have a shared LM head for all transformer layers in the model. This makes training faster, require less memory consumption for both training and inference, and eases deployment and maintenance. Hence, as shown in Figure~\ref{fig:motivation}, we train a deep learning model that is equivalent to an ensemble of models of various depths, capable of skipping from different transfomer layers to the LM head.

\subsection{Correcting if we Exit Too Early}
\label{sec:motivation:self_speculative}

Regardless if we use heuristics or predictors (as ~\citet{calm,prediction_saturation}) to exit early, or if we modify the training procedure to make models predict early (as~\citet{DepthAdaptive,SCAN} and this paper as well), it is likely that exiting early during inference will lead to a reduction in accuracy. It will be ideal if there is a way to verify if an early prediction is accurate, and correct it by executing remaining layers. Some approaches like~\citet{SCAN} proposed a confidence heuristic to decide after executing an early exit if the remaining layers are needed. Here, we leverage speculative decoding techniques to verify the early exit prediction and correct it. Speculative decoding benefits from the fact that verifying the prediction of a group of tokens is faster than generating each token auto-regressively. Hence, we present a \textit{self-speculative decoding} approach where we use early exit to generate each token auto-regressively, and use the remaining layers to verify a group of tokens in parallel, and correct them.

\section{Related Work}
\label{sec:related_work}

\textbf{Dropout}: Dropout was first introduced by~\citet{GeoffHintonDropout} and involved stochastically replacing a portion of output elements of fully-connected layers with zeros during training. We refer to this variant of dropout as \textit{unstructured dropout}. It presented a regularization effect for training, with the purpose of reducing over-fitting. Unstructured dropout was commonly used in convolutional neural networks (CNNs) before batch normalization~\citet{batchnorm} replaced it as a means to improve generalization. However, the introduction of transformers brought it back to light as~\citet{AttentionIsAllYouNeedVaswani} used a dropout rate of 0.1. However, dropout faded again when pretraining dataset sizes increased, e.g., large scale models like Llama~\citet{llama1} and GPT3~\citet{GPT3} do not mention dropout in their papers, and PaLM~\citet{chowdhery2022palm} used a dropout of 0 for pre-training on its large corpus but 0.1 for fine-tuning on small datasets.

The intuition behind dropout’s regularization effect in~\citet{GeoffHintonDropout} was that it enabled training to learn across an ensemble of many models, and avoiding co-adaptation between the model’s nodes. Another intuition is that dropout induces multiplicative noise into node’s activations~\citet{Goodfellow-et-al-2016}, and training a model with noise makes it more robust to the distribution shift between training and validation data. ~\citet{liu2023dropout_reduces_underfitting} showed that dropout induces noise to mini-batch gradients and makes them more aligned to the full dataset gradient in the early epochs of training, and hence showed that applying dropout either early in training or late in training is better than applying dropout throughout training.

\textbf{Layer Dropout}: Skipping layers stochastically during training is referred to in literature with different terms such as \textit{stochastic depth} or \textit{layer dropout}. It was first explored in ResNets by~\citet{stochastic_depth}. ConvNext~\citet{liu2022convnext} used higher layer dropout rates for larger models: 0.1/0.4/0.5/0.5 for ConvNeXt-T/S/B/L respectively when trained on ImageNet~\citet{DenDon09Imagenet}. However, when training on the larger ImageNet-22K dataset, ConvNeXt used smaller layer dropout rates: 0.0/0.0/0.1/0.1/0.2. In language models, LayerDrop~\citet{LayerDrop} applied dropout to every other transformer layer, which increased its robustness to pruning layers at inference time. ~\citet{PLD} increased the pretraining speed of BERT by applying a dropout rate that progressively increased every iteration as well as every layer. To the best of our knowledge, layer dropout for training decoder-only models, or scaling language models to large model sizes or large datasets has not been explored. Moreover, our paper is the first to propose using layer dropout to improve early exit inference.


\textbf{Early Exit}: Exiting early in deep learning has first been explored in CNNs~\citet{panda2016conditional,teerapittayanon2017branchynet}. They added branch modules at different exit points in a deep learning network and introduced additional loss functions during training to improve the accuracies of those early exits. In language models, early exit was explored in encoder-only models by BERxiT~\citet{xin-etal-2021-berxit} and BE3R~\citet{BE3R}, in encoder-decoder models by ~\citet{DepthAdaptive} and CALM~\citet{calm}, and in decoder-only models by SkipDecode~\citet{prediction_saturation,delcorro2023skipdecode}.

~\citet{DepthAdaptive} added a dedicated LM head for each decoder layer in an encoder-decoder translation model, and explored different early exit granularities (per token as well as per sequence) in both training and inference, and explored exiting early both statically and dynamically, using auxiliary layers that learned during training when to exit. CALM~\citet{calm} built upon ~\citet{DepthAdaptive} and started with a model pretrained with early exit losses, and focused on finding optimal criteria to decide which layer to exit at during inference. SkipDecode~\citet{delcorro2023skipdecode} fine-tuned a decoder-only model to skip more layers for later tokens within a sequence. ~\citet{jumpingToConclusion} started with pretrained models and finetuned auxiliary fully-connected layers to map the embeddings outputted by earlier layers to later layers. ~\citet{zeng2023learningskiplanguagemodeling} and ~\citet{raposo2024mixtureofdepthsdynamicallyallocatingcompute} added trainable routers to determine which layers to skip during inference during pretraining. In our proposed solution, we do not introduce any additional modules or linear layers for early exit, and instead used a shared exit for all layers.

\textbf{Speculative Decoding}: Speculative decoding~\citet{yaniv_speculative_decoding,Chen2023AcceleratingLLSpeculativeSampling} is a popular acceleration technique for language models. It is based on the fact that auto-regressive decoding of decoder models are slow as they generate one token a time, while measuring the likelihood of a group of generated tokens in parallel is faster. It uses a fast, less accurate model, referred to as the \textit{draft} model, to generate multiple tokens auto-regressively, and a large, slower, more accurate \textit{main} model, to verify the tokens in parallel, and correct them when needed. The draft model could have the same or different architecture as the main model, or could be a compressed version of the model. ~\citet{zhang2023draft} recently proposed a self-speculative decoding approach where the draft model is the same as the main model, but with a group of intermediate attention and feed forward network (FFN) layers skipped. The advantage of our proposed solution compared to~\citet{zhang2023draft} is that verification and correction stages can reuse the activation and KV cache from the draft stage as both stages execute the same early layers in the same order, while ~\citet{zhang2023draft} can not reuse them as it skips intermediate layers. ~\citet{hooper2024speed} used shared transformer layer groups and a shared LM head to exit each token at a different layer and execute different layer groups in a pipeline fashion.

\section{Proposed Solution}
\label{sec:proposed_solution}

Our approach has three different stages: 

\begin{enumerate}[noitemsep, nolistsep, topsep=0pt]
    \item Training using Layer Dropout \& Early Exit Loss
    \item Inference using Early Exit 
    \item Verification and Correction using Speculative Decoding
\end{enumerate}

We explain each stage in the following sub-sections.

\subsection{Training using Layer Dropout \& Early Exit Loss}
\label{sec:proposed_solution:training}


We denote the input tokens to a transformer model as $X$ and its output as $Y$, with an embedding layer that maps the token indices to token embeddings, $x_{0}$, and a transformer model with $L$ transformer layers, where transformer layer $l$ evolves embeddings outputted from its previous layer, $x_{l+1} = x_l + f_l(x_l)$, and a final LM head that maps the embedding outputs of the last layer, $x_{L}$ to logits, $e_L=g(x_L)$. We denote the cross entropy loss function that is usually used to train language models as $J_{\text{CE}}(e_L,Y)$.

\subsubsection{Layer Dropout}
\label{sec:proposed_solution:training:layer_dropout}

The first modification we apply to common training recipes, is to apply layer dropout. Hence the transformer layer operation at layer $l$ and training iteration $t$ changes to:
\begin{equation}
    x_{l+1,t} = x_{l,t} + M(p_{l,t}) f_l(x_{l,t})
\end{equation}
where $p_{l,t}$ is the dropout rate of layer $l$ at iteration $t$, $M(p)$ is a Bernoulli function that returns 0 with probability $p$ and returns 1 with probability $1-p$. We apply the dropout operation on each sample separately within a batch. We remove the dropped samples from a batch, apply the transformer operation $f_l$ on the remaining samples, and then concatenate the output with the dropped samples. To ensure higher speedup during training, we seed the random number generator for each GPU with the same seed, so that each transformer layer at each iteration will drop the same number of samples. 

The dropout rate can be different at each layer $l$ and training iteration $t$, $p_{l,t}$: 
\begin{equation}
    p_{l,t} = S(t)D(l)p_{max}
\end{equation}
where $p_{max}$ is a hyperparameter that sets the maximum dropout rate in the model during training, $D(l)$ is a per-layer scaling function, and $S(t)$ is a per-time step scaling function. We found that the best per-layer scaling is to increase dropout rate exponentially across layers from 0.0 in layer 0, to 1.0 in last layer, $L-1$:
\begin{equation}
    D(l) = e^{\frac{l\text{ln}2}{L-1}} - 1
\end{equation}
For scaling across time, $S(t)$, we found that if we start with a pre-trained model and perform continual pre-training or finetuning, it is best to not scale across time and hence set $S(t)=1$. However, for pretraining from scratch, we found that an exponential curriculum, $S_{exp}(t)$, lead to best accuracies for $T$ training steps:
\begin{equation}
    S_{exp}(t) = e^{\frac{t\text{ln}2}{T-1}} - 1
\end{equation}

\subsubsection{Early Exit Loss}
\label{sec:proposed_solution:training:early_exit}

To boost prediction accuracy of lower layers, we need to ensure that the model's LM head, $g$, is capable of unembedding outputs of different layers. Hence, during training, we augment layer dropout with early exit loss at each layer. During training we supervise the model directly to connect the early exit layers to the LM head, this enables us to directly supervise the lower layers for the language modeling task. The total loss of the model at iteration $t$ is:
\begin{equation}
    J(X,Y,t) = \sum_{l=0}^{l=L-1} \tilde{e}(t,l)J_{\text{CE}}(g(x_{l+1}),Y) 
\end{equation}
Where $\tilde{e}(t,l)$ is a normalized per-layer loss scale, whose sum across all layers is equal to 1:
\begin{equation}
    \tilde{e}(t,l) = \frac{C(t,l)e(l)}{\sum_{i=0}^{i=L-1}C(t,i)e(i)}
\end{equation}

$C(t,l)$ is a binary curriculum function that determines if we enable early exit of layer $l$ at iteration $t$. We build upon ~\citet{DepthAdaptive} and set a scale that increases across layers, such as the scale at one layer is proportional to the sum of the scales of all previous layers:
\[
    e(l)= 
\begin{cases}
    e_{scale}\sum_{i=0}^{i=l}i,             & \text{if } 0 \leq l < L-1\\
    L-1+e_{scale}\sum_{i=0}^{i=L-2}i,         & \text{if } l = L-1\\
\end{cases}
\]

This way, we penalize later layers with quadratically higher weight, as predicting in later layers is easier. $0 \leq e_{scale} \leq 1$ is a hyperparameter that controls the scale of early exit loss.

Note that we do not add additional LM heads as proposed in other early exit papers~\citet{DepthAdaptive,calm}, as we essentially use the same LM head for all layers.

\textbf{Early Exit Loss Curriculum}: We find that adding early exit loss of all layers at all iterations during training slows down training and reduces the accuracy of the last layer. To overcome this, we introduce a curriculum, $C(t,l)$. We have explored 2 different curricula. First, we explored a \textit{rotational early exit curriculum}, $C_{\text{rot},R}$, where we enable early exit at every $R$ layers, and perform circular rotation at each iteration. This way, the early exit at each layer is enabled once every $R$ iterations. Hence, at each training iteration, only $\lceil L/R \rceil$ unembedding operations are applied. Second, we explored a \textit{gradual early exit curriculum}, $C_{\text{grad}}$, where we gradually enable early exit loss from layers $L-1$ to $0$, one layer at a time every $T/2L$ iterations.

Overall, to summarize, the hyperparameters of our training recipe:
\begin{itemize}
    \item Layer Dropout:
    \begin{itemize}
        \item $p_{max}$: maximum dropout rate of last layer of the model,
        \item $S(t)$: layer dropout curriculum. We use either no curriculum $S(t)=1$ for finetuning or continual pretraining, or an exponential curriculum, $S(t)=S_{exp}(t)$ for pretraining from scratch,
    \end{itemize}
    \item Early Exit Loss:
    \begin{itemize}
        \item $e_{scale}$: scalar scale of loss of earlier layers,
        \item $C(t,l)$: early exit loss curriculum, either rotational, $C_{\text{rot},R}(t,l)$, or gradual, $C_{\text{grad}}(t,l)$ 
        \begin{itemize}
            \item $R$: is a dilation across layers for rotational early exit loss curriculum
        \end{itemize}
    \end{itemize}
\end{itemize}

\subsection{Inference using Early Exit}
\label{sec:proposed_solution:early_exit_inference}

\begin{figure*}[t]
    \vskip 0.2in
    \begin{center}
    \centerline{\includegraphics[width=0.5\columnwidth]{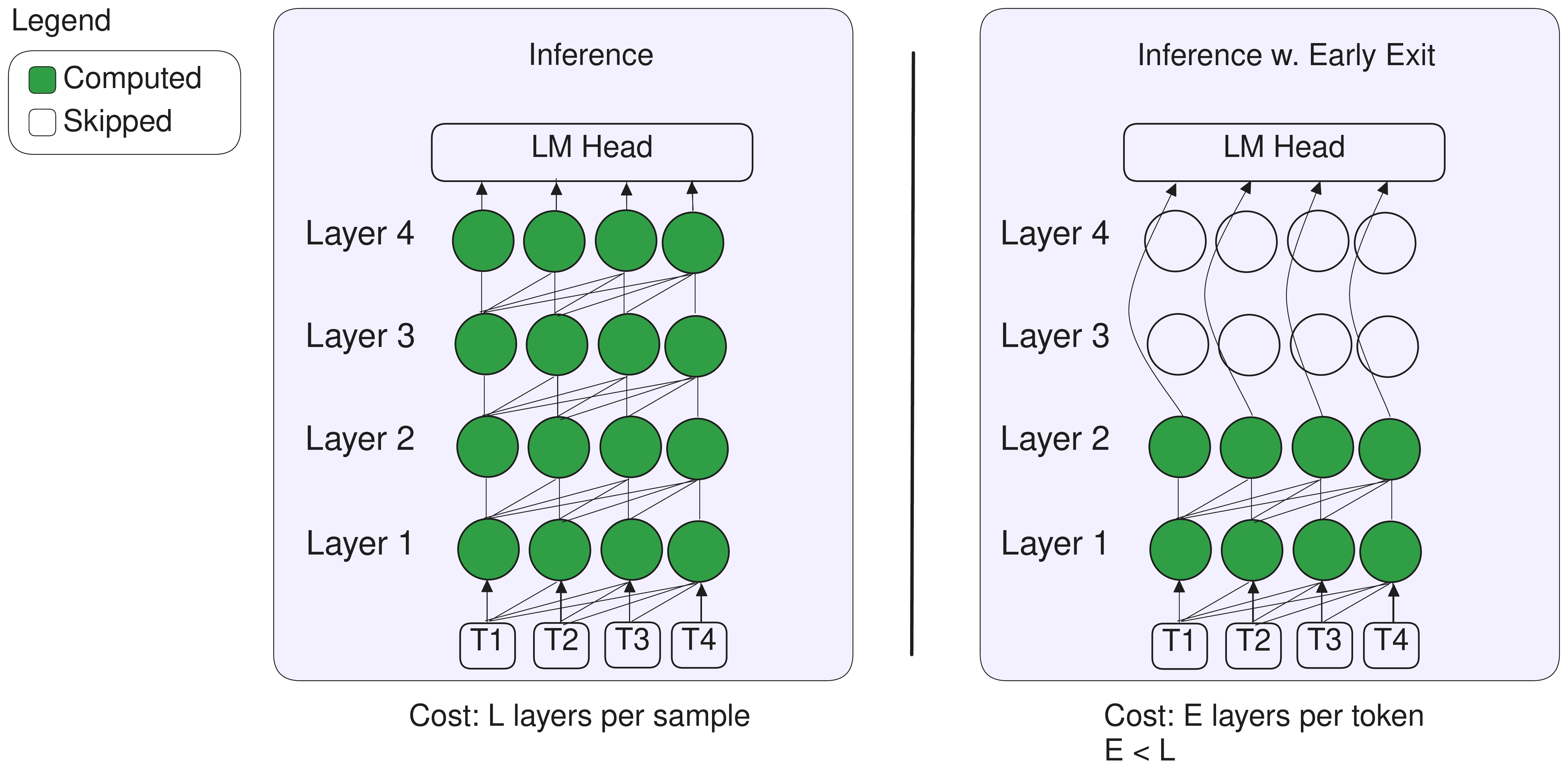}}
    \caption{Early exit inference.}
    \label{fig:early_exit_inference}
    \end{center}
    \vskip -0.2in
\end{figure*}

When generating each token during autoregressive decoding, we run the first $E$ transformer layers in a model, and skip to the model's LM head, i.e., the model's final output becomes $g(x_{E})$. We explore with different values of $E$ and provide the accuracies in the Results section. 

\subsection{Inference using Self-Speculative Decoding}
\label{sec:proposed_solution:self_speculative_decoding}

\begin{figure*}[!ht]
    \vskip 0.2in
    \begin{center}
    \centerline{\includegraphics[width=0.5\columnwidth]{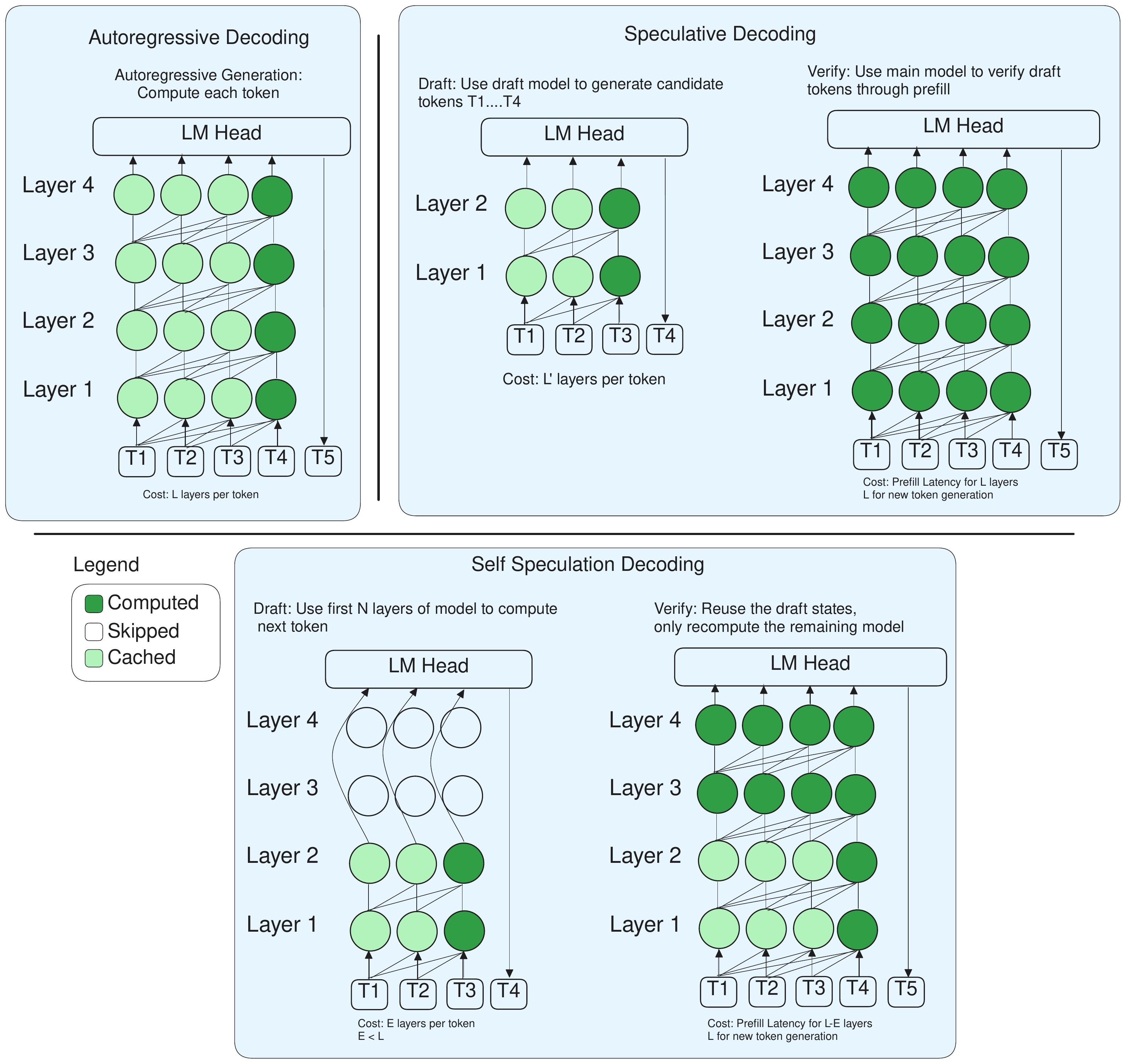}}
    \caption{Comparison between autoregressive decoding, speculative decoding, and our proposed self-speculative decoding.}
    \label{fig:self_speculative_decoding}
    \end{center}
    \vskip -0.2in
\end{figure*}

With layer dropout and early exit loss in training, we show it is possible to speedup autoregressive generation by exiting early, but this comes at an accuracy cost compared to using the full model. Speculative decoding \citet{yaniv_speculative_decoding,Chen2023AcceleratingLLSpeculativeSampling} is able to leverage a faster yet less accurate model to speedup generation without accuracy cost. However, this requires storing and training 2 models, increasing memory and training requirements.

We introduce a novel self-speculative decoding algorithm built on top of early exit, enabling us to reduce memory through the use of a single model and latency of traditional speculative decoding through re-using hidden states in draft and verify steps. As shown in Figure~\ref{fig:self_speculative_decoding}, our self-speculation algorithm consists of 2 key steps (1) \textit{Self-Drafting}, using the early exit to draft tokens from the same model (2) \textit{Self-Verification}, using the remaining layers to validate the prediction. To enable re-use in (1) and (2), we develop a novel \textit{Cache Reuse} technique that unifies the KV cache and storing the exit query. We provide a high level description of the algorithm in sections \S \ref{sec:proposed_solution:self_speculative_decoding:self_drafting} and \ref{sec:proposed_solution:self_speculative_decoding:self_verification} and provide pseudo code in \ref{sec:appendix:speculation_algo}.

\subsubsection{Self-Drafting}
\label{sec:proposed_solution:self_speculative_decoding:self_drafting}
The first step in speculative decoding is to define a set of draft tokens $D_{0...d-1}$. In our algorithm, we compute the first $d$ draft tokens through early exit. We refer to $d$ as the number of speculations. We leverage a subset of the LLM and conduct auto-regressive inference exiting at layer $E$.

Our training recipe enabled us to train the model once to get an ensemble of different candidate draft models at each layer depth. We can evaluate exiting at different layers and observe a trade off between latency and accuracy.

\subsubsection{Self-Verification}
\label{sec:proposed_solution:self_speculative_decoding:self_verification}

The next step in speculative decoding is verification. Verification leverages the full LLM to predict the next token for each draft token in a single forward pass. We then assess to see where the draft tokens and verified tokens agree. All the draft tokens up till the disagreement point are added to the output along with the next verified token and the process continues from the draft stage.

In our self-speculative decoding algorithm, the self-verification stage critically only requires computing the remaining layers of the model that were not used in the draft stage. For a model with $L$ layers, the number of verification layers is $L - E$. In order to re-use the first $E$ layers from the draft stage we employ some modifications to the KV cache as we show in the subsequent subsection.

\subsubsection{Reusing the Cache}
\label{sec:proposed_solution:self_speculative_decoding:resuing_cache}

In autoregressive transformers, the KV cache is a critical component of efficient generation, allowing us to avoid recomputing prior KV pairs in each transformer layer. 

As our draft stage uses the first $E$ layers of the model and the verification stage uses the remaining $L - E$ layers, we are able to re-use a significant amount of compute between the 2 stages:
\begin{itemize}[noitemsep, nolistsep, topsep=0pt]
    \item \textbf{Single KV Cache}  As the draft model and verification model operate on the same model using the same order of layers, the first $E$ layers are shared in both steps. Hence, in the draft stage, the KV cache in the first $E$ layers are already computed, so we are able to effectively maintain a single KV cache for the draft and verify steps, reducing memory and latency.
    \item \textbf{Exit Query Cache}: To further reduce computation of the first $E$ layers, we introduce an \textit{exit query cache} that saves the query vector of exit layer $E-1$ for verification to directly continue from layer $E$ to last layer $L$. Critically note that we need to save only the query for the exit layer. We term the union of the KV cache and the exit query as KVQ cache.
\end{itemize}





\section{Experiments}
\label{sec:experiments}


We would like to evaluate our training recipe on different types of training, whether pretraining from scratch or finetuning. To verify our approach, we run different types of training experiments:
\begin{itemize}[noitemsep, nolistsep, topsep=0pt]
    \item \textbf{Continual Pretraining}: start with a pretrained model and continue pretraining on 52B tokens from a corpus of diverse data containing natural language text and code. We experiment using pretrained Llama2 7B (32 layers), with $p_{\text{max}}=0.1$, $e_{scale}=0.2$, $C_{\text{rot},R=8}$, and Llama2 13B (40 layers), with $p_{\text{max}}=0.1$, $e_{scale}=0.1$, $C_{\text{rot},R=39}$. We later continually pretrained Llama3 8B on 419B tokens with $p_{\text{max}}=0.1$, $e_{scale}=0.1$, $C_{\text{rot},R=8}$, and and Llama3.2 1B on 839B tokens with $p_{\text{max}}=0.1$, $e_{scale}=0.1$, $C_{\text{rot},R=8}$.
    \item \textbf{Pretraining from Scratch}: start with randomly initialized model and pretrain on 26B tokens from a corpus of diverse data containing natural language text and code. We experiment with Llama2 1.5B (a custom small Llama-like model with 24 layers) (see ~\ref{sec:appendix:experiment_details:architectures} for architecture details) with $p_{\text{max}}=0.1$, $e_{scale}=0.2$, $C_{\text{rot},R=23}$ and Llama2 7B (32 layers) with $p_{\text{max}}=0.2$, $e_{scale}=0.2$, $C_{\text{rot},R=31}$. Following ~\citet{GeoffHintonDropout} we use higher learning rates when layer dropout is greater than 0.0.
    \item \textbf{Finetuning on Code Data}: start with pretrained Llama1 7B model~\citet{llama1} and finetune on 5.2B tokens of CodeLlama~\citet{codellama} data mix. We use $p_{\text{max}}=0.1$, $e_{scale}=1.0$, $C_{\text{rot},R=16}$.
    \item \textbf{Finetuning on Task-Specific Dataset}: start with a pretrained Llama 1.5B (24 layers) and finetune on TOPv2~\citet{TOPv2}, a multi-domain task-oriented compositional semantic parsing dataset. We post processed the dataset into a JSON format to be more aligned with code pre-training. We report our results on the TOPv2 evaluation set. We use $p_{\text{max}}=0.2$, $e_{scale}=1.0$, $C_{\text{grad}}$.
\end{itemize}
We try different variants of \System: layer dropout only (LD), early exit loss only (EE), and both layer dropout and early exit loss (LD+EE). We provide more details about training hyperparameters in Appendix~\ref{sec:appendix:experiment_details}.

\section{Results}
\label{sec:results}

\subsection{Early Exit Inference Results}
\label{sec:results:early_exit}
After training each model configuration, we evaluate accuracy of exiting early at different layers. 

\begin{figure*}[!ht]
     \centering
     \begin{subfigure}[b]{0.48\textwidth}
         \centering
         \includegraphics[width=\columnwidth]{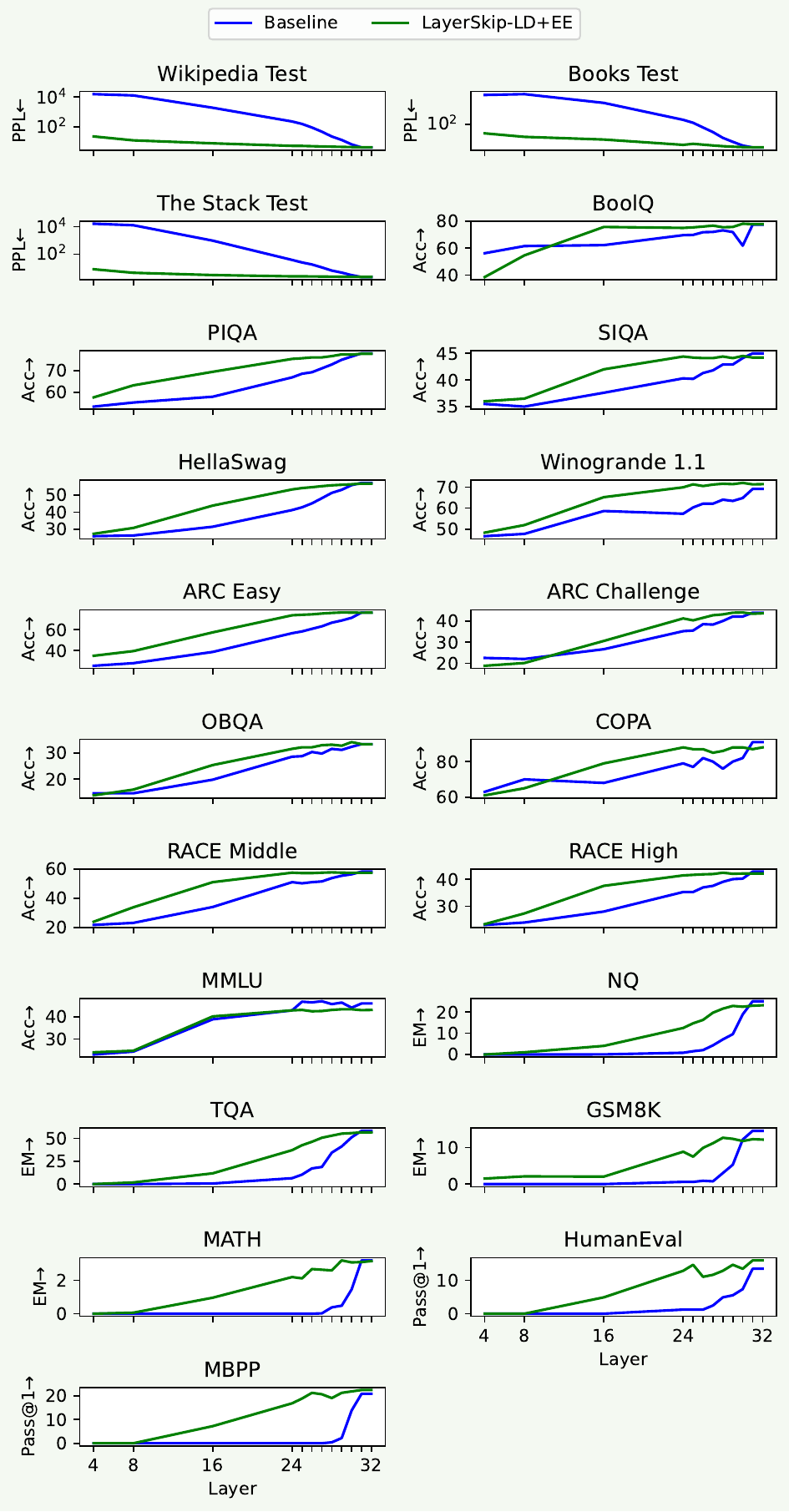}
         \caption{Llama2 7B}
         \label{fig:early_exit:continual_pretraining:llama2_7B}
     \end{subfigure}
     \hfill
     \begin{subfigure}[b]{0.48\textwidth}
         \centering
         \includegraphics[width=\columnwidth]{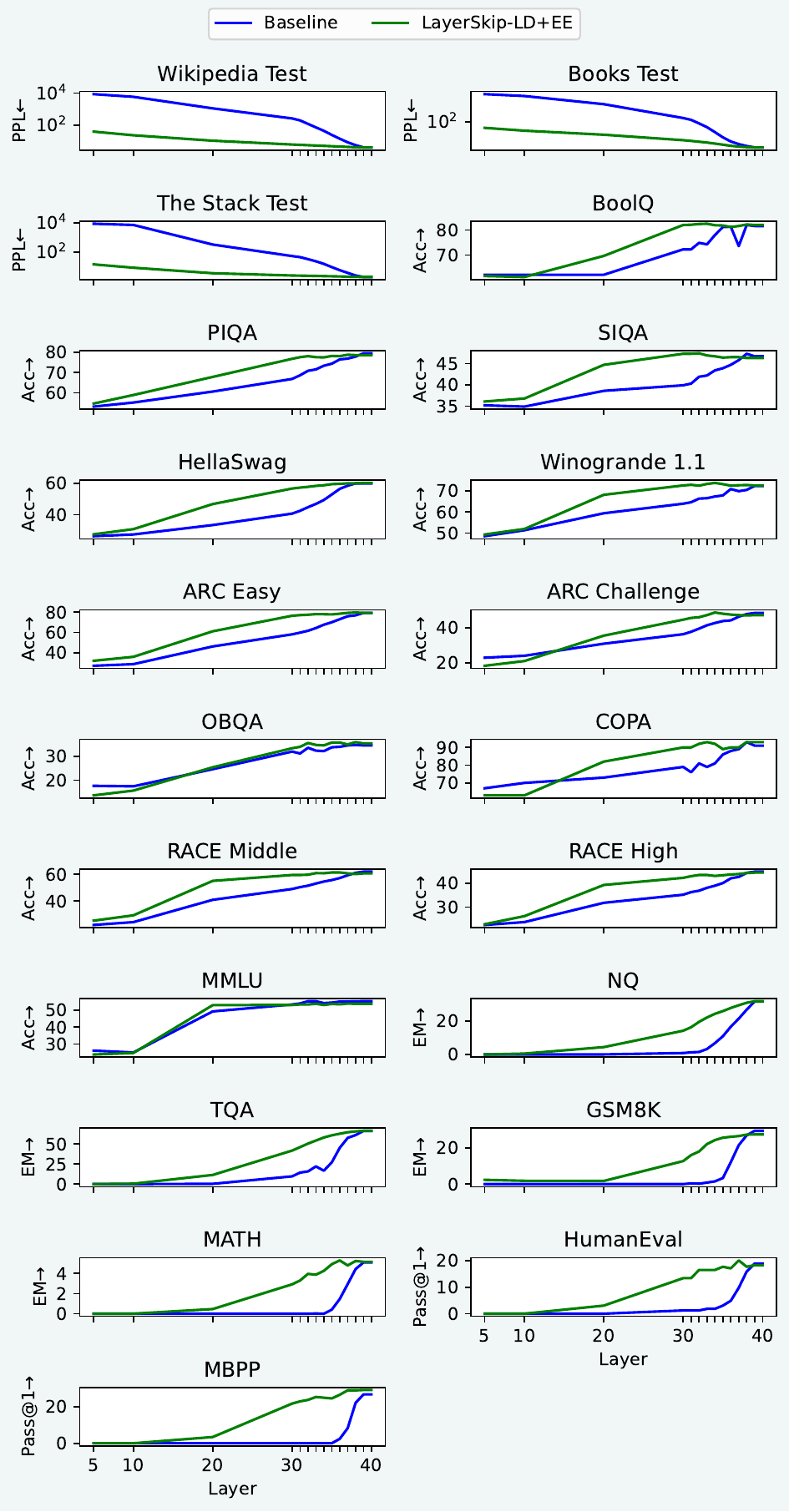}
         \caption{Llama2 13B}
         \label{fig:early_exit:continual_pretraining:llama2_13B}
     \end{subfigure}
     \caption{Early exit evaluation of continual pretraining.}
     \label{fig:early_exit:continual_pretraining}
\end{figure*}

\begin{figure*}[!ht]
     \centering
     \begin{subfigure}[b]{0.75\textwidth}
         \centering
         \includegraphics[width=\columnwidth]{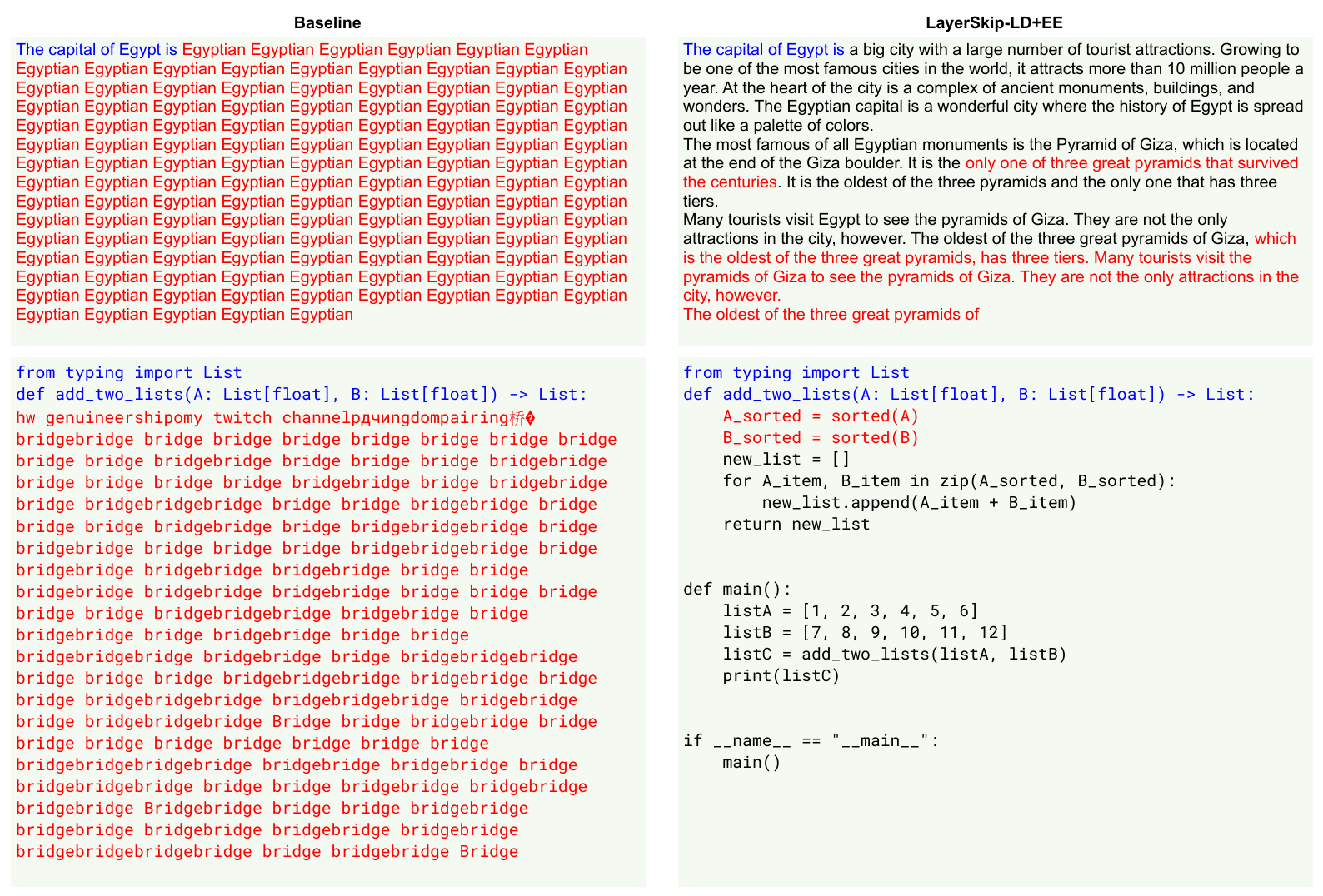}
         \caption{Llama2 7B (that has 32 layers) exiting at layer 20.}
         \label{fig:early_exit:generation_continual_pretraining:llama2_7B}
     \end{subfigure}
     \hfill
     \begin{subfigure}[b]{0.75\textwidth}
         \centering
         \includegraphics[width=\columnwidth]{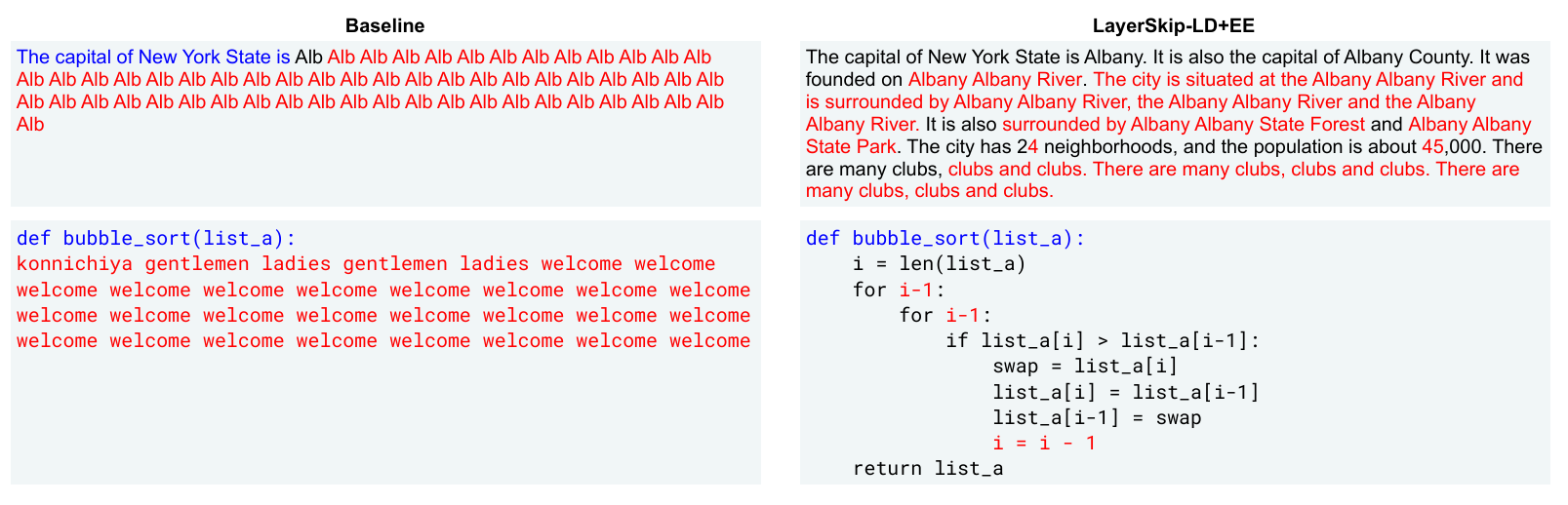}
         \caption{Llama2 13B (that has 40 layers) exiting at layer 24.}
         \label{fig:early_exit:generation_continual_pretraining:llama2_13B}
     \end{subfigure}
     \caption{Early exit text generation examples for models continually pretrained with LayerSkip. \textcolor{blue}{Blue}: The prompt fed into the model. \textcolor{red}{Red}: incorrect phrases or words generated by the model (whether factually or grammatically wrong, or hallucinations). With self-speculative decoding, we fix those incorrect phrases by verifying with remaining layers.}
     \label{fig:early_exit:generation_continual_pretraining}
\end{figure*}

\textbf{Continual Pretraining} In Figure~\ref{fig:early_exit:continual_pretraining}, we present our results for Llama2 7B and 13B on a diverse set of evaluation tasks (see \S~\ref{sec:appendix:experiment_details:evaluation_tasks} for task details) and compare with the baseline model from ~\citet{llama2}. In Table~\ref{tab:early_exit:continual_pretraining} we zoom in and show the specific values of accuracies for the last layer and middle layer of each model. In Figure~\ref{fig:early_exit:generation_continual_pretraining} we show sample text generations for exiting at earlier layers for both models with and without continual pretraining with \System. Overall, for earlier layers, \System\ is clearly better than the baseline. For last layer accuracy, \System\ has minimal drop in accuracy compared to baseline.

It is noteworthy that some ``classification'' tasks, i.e., multiple choice question or true/false question tasks, maintain relatively decent accuracy on earlier layers on the baseline model, while open-ended ``generation'' tasks drop drastically. Surprisingly, MMLU~\citet{MMLU} which is considered a challenging task, only drops from 55.2\% to 49.2\% on Llama2 13B baseline from the last to the middle layer. This could be because classification tasks are evaluated on generating one token only while generation tasks are evaluated on the accuracy of many tokens, and an error in one token may have a compounding effect when generating later tokens. Moreover, classification tasks evaluate a token out of 4 or 2 possible outcomes, while generation tasks evaluate each token out of thousands of possible entries in the LLM's dictionary. We observe \System's significant importance on generation tasks, e.g., NaturalQuestions~\citet{NQ} drops from 25.1\% to 0\% when exiting in middle layers of Llama2 7B, but jump to 4\% when using \System. 

\begin{table*}[!ht]
\centering
\begin{NiceTabular}{lcccc|cccc}
\CodeBefore
\rectanglecolor{greenlight}{1-2}{1-5}
\rectanglecolor{greenlight}{2-2}{40-3}
\rectanglecolor{greenlight}{2-4}{40-5}
\rectanglecolor{turquoiselight}{1-6}{1-9}
\rectanglecolor{turquoiselight}{2-6}{40-7}
\rectanglecolor{turquoiselight}{2-8}{40-9}
\rectanglecolor{white}{5-1}{5-9}
\rectanglecolor{white}{9-1}{10-9}
\rectanglecolor{white}{20-1}{21-9}
\rectanglecolor{white}{24-1}{25-9}
\rectanglecolor{white}{27-1}{28-9}
\rectanglecolor{white}{31-1}{32-9}
\rectanglecolor{white}{35-1}{36-9}
\Body
\toprule
                                                                            & \multicolumn{4}{c}{\textbf{Llama2 7B}}                                                              & \multicolumn{4}{c}{\textbf{Llama2 13B}} \\
\toprule
                                                                            & \multicolumn{2}{c}{Last Layer}            & \multicolumn{2}{c}{Middle Layer}               & \multicolumn{2}{c}{Last Layer}            & \multicolumn{2}{c}{Middle Layer} \\
                                                                            & \multicolumn{2}{c}{(Layer 32)}            & \multicolumn{2}{c}{(Layer 16)} & \multicolumn{2}{c}{(Layer 40)}            & \multicolumn{2}{c}{(Layer 20)}   \\
\cmidrule{2-9}
                                                                            & Baseline              & LayerSkip         & Baseline                & LayerSkip & Baseline              & LayerSkip         & Baseline                & LayerSkip \\
\midrule
\multicolumn{6}{l}{\textbf{Eval Perplexity} ↓} \\
\midrule
Wikipedia                                                                   & 4.32	& \textbf{4.3}	& 1900	& \textbf{8.12} & \textbf{3.97}         & 3.98              & 507                     & \textbf{10.5}      \\
Selected Books                                                              & 1.60	& \textbf{1.06}	& 4390 &	\textbf{6.53} & \textbf{1.40}          & \textbf{1.40}              & 1170                    & \textbf{11.9}      \\
The Stack                                                                   & 2.15	& \textbf{2.14}	& 968 &	\textbf{2.99} & \textbf{2.05}         & 2.06              & 65.8                    & \textbf{3.71}      \\
\midrule
\multicolumn{6}{l}{\textbf{Common Sense Reasoning Tasks} ↑} \\
\multicolumn{6}{l}{(Multiple Choice Questions / True False Questions)} \\
\midrule
BoolQ                                                                       & 77.4	& \textbf{77.8} &	62.2 &	\textbf{75.7} & 81.6                  & \textbf{82.0}       & 62.2                    & \textbf{69.7}      \\
PIQA                                                                        & \textbf{78.0}	& 77.9 &	57.9 &	\textbf{69.5} & \textbf{79.3}         & 78.5              & 62.8                    & \textbf{67.8}      \\
SIQA                                                                        & \textbf{44.7}	& 44.2 &	37.8 &	\textbf{42.0} & \textbf{46.7}         & 46.3              & 40.7                    & \textbf{44.7}      \\
HellaSwag                                                                   & \textbf{57.0} &	56.6 &	31.5 &	\textbf{43.8} & 60.1         & \textbf{60.3}     & 35.6                    & \textbf{46.8}      \\
WinoGrande                                                                  & 69.8 &	\textbf{71.4} &	58.6 &	\textbf{65.2} & 72.3                  & \textbf{72.5}     & 59.4                    & \textbf{68.1}        \\
ARC-e                                                                       & \textbf{76.5}	& \textbf{76.5} &	38.6 &	\textbf{57.5} & \textbf{79.4}         & 79.2              & 48.8                    & \textbf{61.1}      \\
ARC-c                                                                       & \textbf{43.8}	& 43.6 &	26.8 &	\textbf{30.6} & \textbf{48.3}         & 47.3              & 31.9                    & \textbf{35.6}      \\
OBQA                                                                        & \textbf{33.4}	& \textbf{33.4} &	19.6 &	\textbf{25.4} & 34.4                  & \textbf{35.4}     & 23.8                    & \textbf{25.4}      \\
COPA                                                                        & \textbf{90} &	88 &	68 &	\textbf{79} & 91                    & \textbf{93}       & 73                      & \textbf{82}        \\
\midrule
\multicolumn{6}{l}{\textbf{Reading Comprehension} ↑} \\
\multicolumn{6}{l}{(Multiple Choice Questions)} \\
\midrule
RACE Middle                                                                 &   \textbf{58.2}                    &       57.4            &  34.0                       & \textbf{51.1} & \textbf{62.0} &	60.7 &	40.9 &	\textbf{55.1}          \\
RACE High                                                                   &   \textbf{42.9}                    &        42.2           &  28.0                       & \textbf{37.6} & \textbf{44.9} &	44.5	& 31.8 &	\textbf{39.3}          \\
\midrule
\multicolumn{6}{l}{\textbf{MMLU} ↑} \\
\multicolumn{6}{l}{(Multiple Choice Questions)} \\
\midrule
MMLU                                                                 &   \textbf{46.0}                    &       43.1            &  38.9                       & \textbf{40.2} & \textbf{55.2} &	53.7 &	49.2 &	\textbf{52.9}          \\
\midrule
\multicolumn{6}{l}{\textbf{Question Answering} ↑} \\
\multicolumn{6}{l}{(Open Ended Answers)} \\
\midrule
NaturalQuestions                                                            & \textbf{25.1} &	23.2 &	0.0554 &	\textbf{4.07} & 31.5         & \textbf{31.8}              & 0.609                   & \textbf{4.43}      \\
TriviaQA                                                                    & \textbf{58.5}	& 56.8 &	0.619 &	\textbf{11.8} & 66.2         & \textbf{66.3}              & 4.36                    & \textbf{11.4}      \\
\midrule
\multicolumn{6}{l}{\textbf{Mathematics} ↑} \\
\multicolumn{6}{l}{(Open Ended Answers)} \\
\midrule
GSM8K                                                                       & \textbf{14.3} &	12.2 &	0 &	\textbf{2.05} & \textbf{29.3}         & 27.4                & 0.0758                  & \textbf{1.74}       \\
MATH                                                                        & \textbf{3.22} &	3.16 &	0 &	\textbf{0.96} & 5.06                  & \textbf{5.16}     & 0                       & \textbf{0.46}      \\
\midrule
\multicolumn{6}{l}{\textbf{Code Generation} ↑} \\
\multicolumn{6}{l}{(Open Ended Answers)} \\
\midrule
HumanEval                                                                   & 13.4	& \textbf{15.9} &	0 &	\textbf{4.88} & \textbf{18.9}         & 18.3              & 0                       & \textbf{3.05}      \\
MBPP                                                                        & 21.0	& \textbf{22.4} &	0 &	\textbf{7.20} & 26.4                  & \textbf{29.0}     & 0                       & \textbf{3.40}       \\
\bottomrule
\end{NiceTabular}
\caption{Evaluation of continual pretraining of Llama2 7B and Llama2 13B.}
\label{tab:early_exit:continual_pretraining}
\end{table*}

For Llama3 8B and Llama 3.2 1B, we notice a larger drop in last layer accuracy when continually pretraining with \System, despite training on more tokens. The perplexity of earlier layers of the baseline Llama3 models are 2 or 3 orders of magnitude higher than their Llama2 counterparts. Reducing such high perplexity of earlier layers while maintaining accuracy of last layer is more challenging and could be the cause of the drop in last layer accuracy. The increase in perplexity of earlier layers of Llama3 compared to Llama2 could be due to the increase of number of pretraining tokens (8T for Llama3 versus 2T for Llama2), as the ablation in Section~\ref{sec:ablation:scaling_pretraining_tokens} confirms. This could be a motivation to consider our \System\ recipe in pretraining future LLMs from scratch.

\begin{table*}[!ht]
\centering
\begin{NiceTabular}{lcccc|cccc}
\CodeBefore
\rectanglecolor{cyanlight}{1-2}{1-5}
\rectanglecolor{cyanlight}{2-2}{40-3}
\rectanglecolor{cyanlight}{2-4}{40-5}
\rectanglecolor{bluelight}{1-6}{1-9}
\rectanglecolor{bluelight}{2-6}{40-7}
\rectanglecolor{bluelight}{2-8}{40-9}
\rectanglecolor{white}{5-1}{5-9}
\rectanglecolor{white}{9-1}{10-9}
\rectanglecolor{white}{20-1}{21-9}
\rectanglecolor{white}{24-1}{25-9}
\rectanglecolor{white}{27-1}{28-9}
\rectanglecolor{white}{31-1}{32-9}
\rectanglecolor{white}{35-1}{36-9}
\Body
\toprule
                                                                            & \multicolumn{4}{c}{\textbf{Llama3 8B}}                                                              & \multicolumn{4}{c}{\textbf{Llama3.2 1B}} \\
\toprule
                                                                            & \multicolumn{2}{c}{Last Layer}            & \multicolumn{2}{c}{Middle Layer}               & \multicolumn{2}{c}{Last Layer}            & \multicolumn{2}{c}{Middle Layer} \\
                                                                            & \multicolumn{2}{c}{(Layer 32)}            & \multicolumn{2}{c}{(Layer 16)} & \multicolumn{2}{c}{(Layer 16)}            & \multicolumn{2}{c}{(Layer 8)}   \\
\cmidrule{2-9}
                                                                            & Baseline              & LayerSkip         & Baseline                & LayerSkip & Baseline              & LayerSkip         & Baseline                & LayerSkip \\
\midrule
\multicolumn{6}{l}{\textbf{Eval Perplexity} ↓} \\
\midrule
Wikipedia                                                                   & 5.63	& \textbf{5.38}	& \num{1.10e05}	& \textbf{12.2}              & 8.52	& \textbf{8.38}	& \num{2.18e05}	& \textbf{22.1}      \\
Selected Books                                                              & \textbf{1.05}	& \textbf{1.05}	& \num{2.69e05} &	\textbf{8.86}            & \textbf{1.23}	& 2.39	& \num{1.86e05} &	\textbf{15.3}      \\
The Stack                                                                   & 2.56	& \textbf{2.37}	& \num{2.95e05} &	\textbf{3.75}            & 3.34	& \textbf{2.95}	& \num{2.71e05} &	\textbf{5.76}      \\
\midrule
\multicolumn{6}{l}{\textbf{Common Sense Reasoning Tasks} ↑} \\
\multicolumn{6}{l}{(Multiple Choice Questions / True False Questions)} \\
\midrule
BoolQ                                                                       & 83.3	& \textbf{84.0} &	62.3 &	\textbf{79.5}                     & \textbf{64.2}	& 63.5 &	\textbf{62.2} &	57.7      \\
PIQA                                                                        & \textbf{79.0}	& \textbf{79.0} &	59.9 &	\textbf{71.3}                    & \textbf{75.0}	& 72.9 &	56.7 &	\textbf{62.8}      \\
SIQA                                                                        & \textbf{45.5}	& 45.0 &	35.7 &	\textbf{42.0}                    & 42.0	& \textbf{42.3} &	33.8 &	\textbf{37.2}      \\
HellaSwag                                                                   & \textbf{60.0} &	59.2 &	28.0 &	\textbf{44.8}                    & \textbf{47.2} &	45.5 &	26.5 &	\textbf{31.7}      \\
WinoGrande                                                                  & \textbf{72.8} &	71.7 &	55.4 &	\textbf{65.8}                    & \textbf{61.3} &	60.2 &	50.0 &	\textbf{51.9}        \\
ARC-e                                                                       & 78.7	& \textbf{79.4} &	33.9 &	\textbf{62.7}                    & \textbf{67.1}	& 66.8 &	28.3 &	\textbf{41.7}      \\
ARC-c                                                                       & \textbf{51.0}	& 48.2 &	27.2 &	\textbf{31.9}                     & \textbf{33.0}	& 32.3 &	\textbf{23.9} &	20.2      \\
OBQA                                                                        & \textbf{34.0}	& 33.6 &	19.0 &	\textbf{21.6}                    & \textbf{28.2}	& 26.6 &	19.6 &	\textbf{18.4}      \\
COPA                                                                        & \textbf{89} &	87 &	58 &	\textbf{81}                  & \textbf{79} &	\textbf{79} &	58 &	\textbf{62}        \\
\midrule
\multicolumn{6}{l}{\textbf{Reading Comprehension} ↑} \\
\multicolumn{6}{l}{(Multiple Choice Questions)} \\
\midrule
RACE Middle                                                                 &   \textbf{62.3}                    &       61.1            &  25.9                       & \textbf{51.5} &   51.6                    &       \textbf{52.5}            &  22.7                       & \textbf{32.1}          \\
RACE High                                                                   &   \textbf{44.9}                    &        43.1           &  24.8                       & \textbf{36.6} &   36.7                    &        \textbf{37.1}           &  33.8                       & \textbf{37.2}          \\
\midrule
\multicolumn{6}{l}{\textbf{MMLU} ↑} \\
\multicolumn{6}{l}{(Multiple Choice Questions)} \\
\midrule
MMLU                                                                 &   \textbf{66.5}                    &       60.5            &  27.6                       & \textbf{38.1} &   \textbf{31.2}                    &       24.8            &  25.8                       & \textbf{25.9}          \\
\midrule

\multicolumn{6}{l}{\textbf{Question Answering} ↑} \\
\multicolumn{6}{l}{(Open Ended Answers)} \\
\midrule
NaturalQuestions                                                            & \textbf{30.3} &	27.5 &	0.00 &	\textbf{4.63}       & \textbf{11.9} &	9.06 &	0.00 &	\textbf{1.08}      \\
TriviaQA                                                                    & \textbf{65.5}	& 62.5 &	0.00 &	\textbf{14.8}       & \textbf{36.1}	& 31.3 &	0.00 &	\textbf{1.95}      \\
\midrule
\multicolumn{6}{l}{\textbf{Mathematics} ↑} \\
\multicolumn{6}{l}{(Open Ended Answers)} \\
\midrule
GSM8K                                                                       & \textbf{54.2} &	45.0 &	0.00 &	\textbf{2.27}       & \textbf{5.84} &	3.79 &	0.00 &	\textbf{2.05}       \\
MATH                                                                        & \textbf{17.3} &	12.3 &	0.00 &	\textbf{0.72}       & 1.70 &	\textbf{1.78} &	0.00 &	\textbf{0.04}      \\
\midrule
\multicolumn{6}{l}{\textbf{Code Generation} ↑} \\
\multicolumn{6}{l}{(Open Ended Answers)} \\
\midrule
HumanEval                                                                   & \textbf{37.8}	& 28.7 &	0.00 &	\textbf{7.32}       & \textbf{17.7}	& 9.15 &	0.00 &	\textbf{1.22}      \\
MBPP                                                                        & \textbf{49.0}	& 40.0 &	0.00 &	\textbf{13.6}       & \textbf{25.8}	& 13.2 &	0.00 &	\textbf{0.00}       \\
\bottomrule
\end{NiceTabular}
\caption{Evaluation of continual pretraining of Llama3 8B and Llama3.2 1B.}
\label{tab:early_exit:continual_pretraining_llama3}
\end{table*}

\begin{figure*}[!ht]
     \centering
     \begin{subfigure}[b]{0.48\textwidth}
         \centering
         \includegraphics[width=\columnwidth]{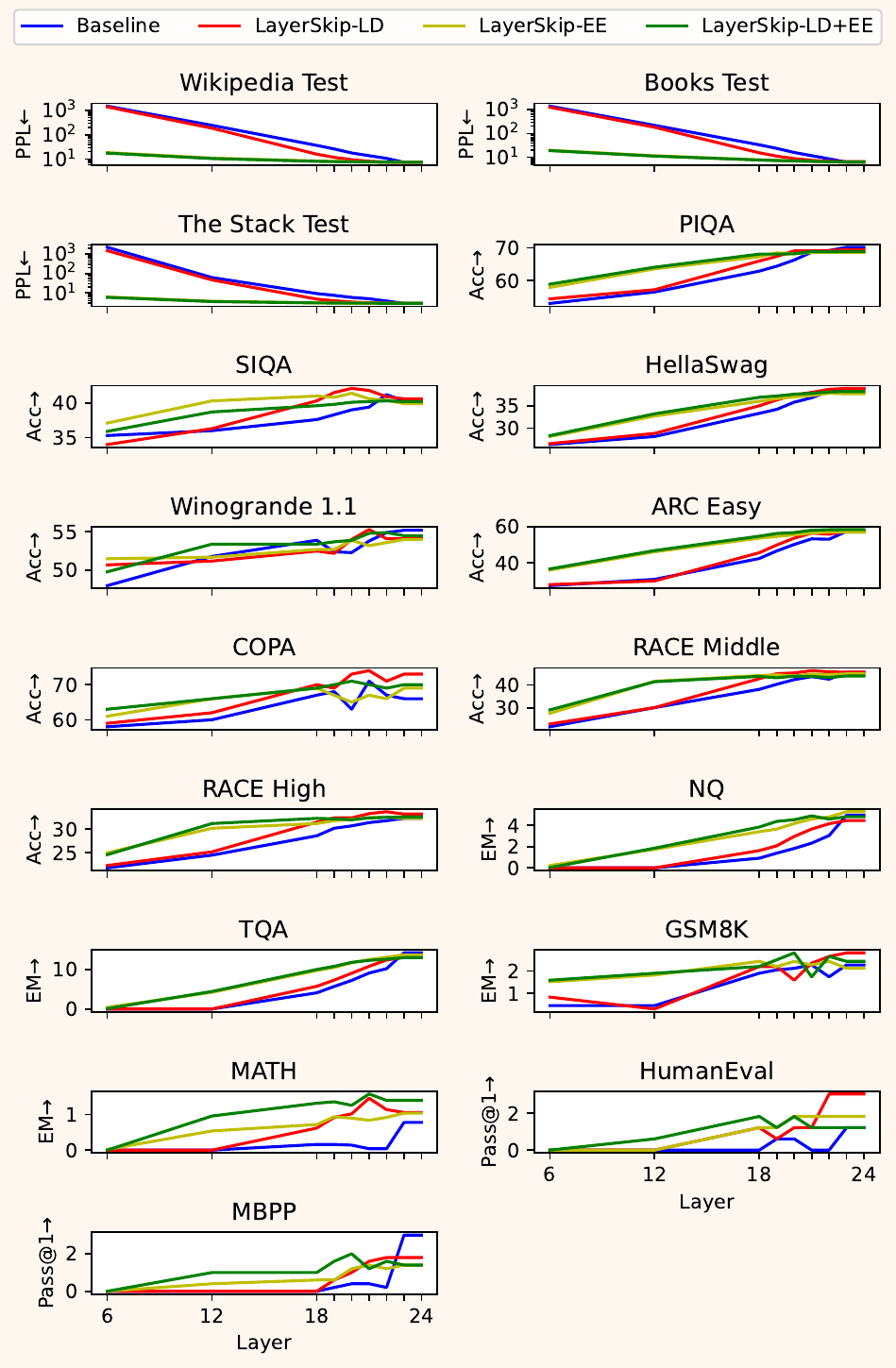}
         \caption{Llama2 1.5B - 26B tokens}
         \label{fig:early_exit:pretraining:llama2_1.5B}
     \end{subfigure}
     \hfill
     \begin{subfigure}[b]{0.48\textwidth}
         \centering
         \includegraphics[width=\columnwidth]{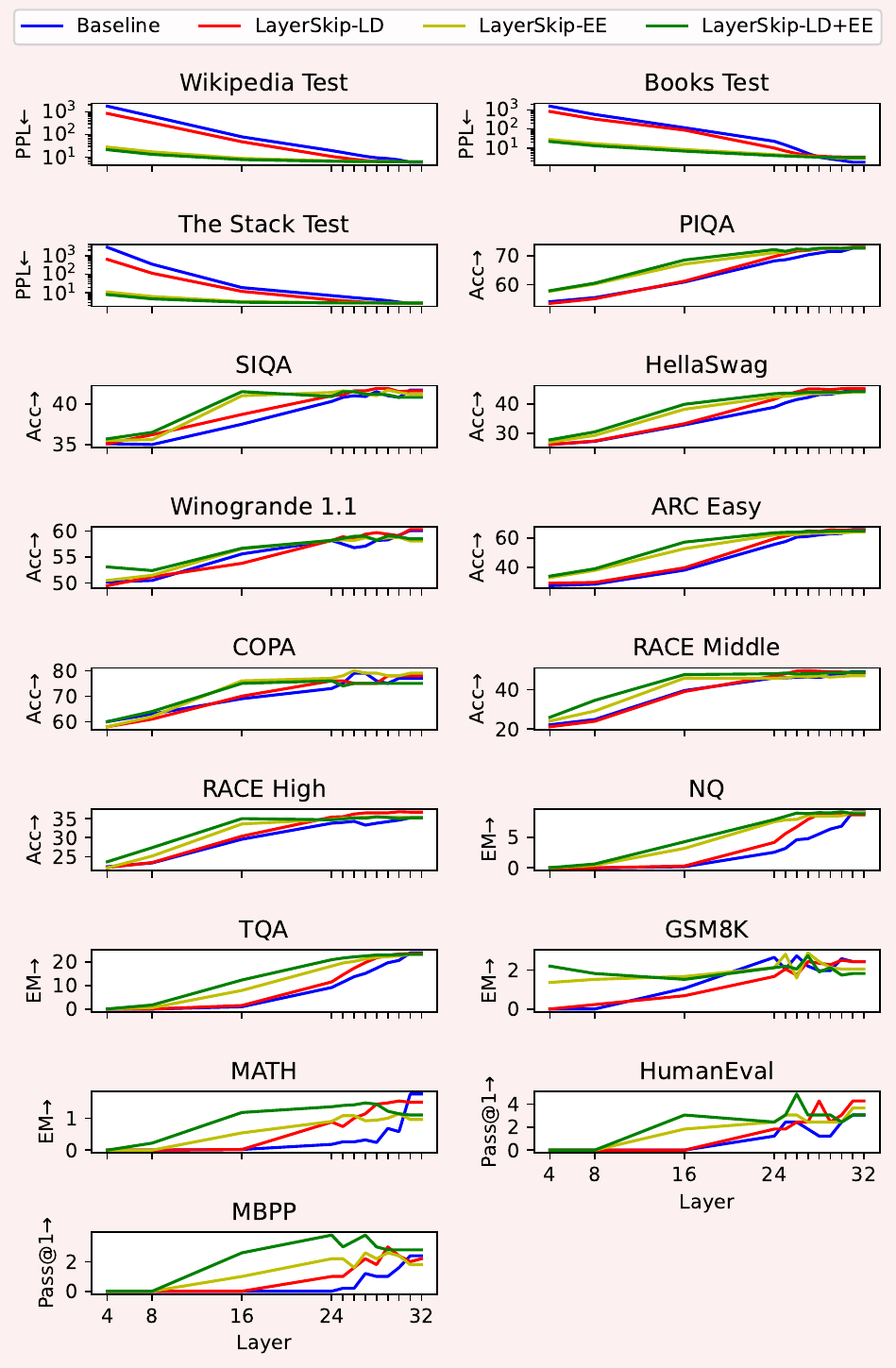}
         \caption{Llama2 7B - 26B tokens}
         \label{fig:early_exit:pretraining:llama2_7B}
     \end{subfigure}
     \caption{Early exit evaluation of pretraining from scratch on 26B tokens.}
     \label{fig:early_exit:pretraining}
\end{figure*}

\begin{figure*}[!ht]
     \centering
     \begin{subfigure}[b]{0.75\textwidth}
         \centering
         \includegraphics[width=\columnwidth]{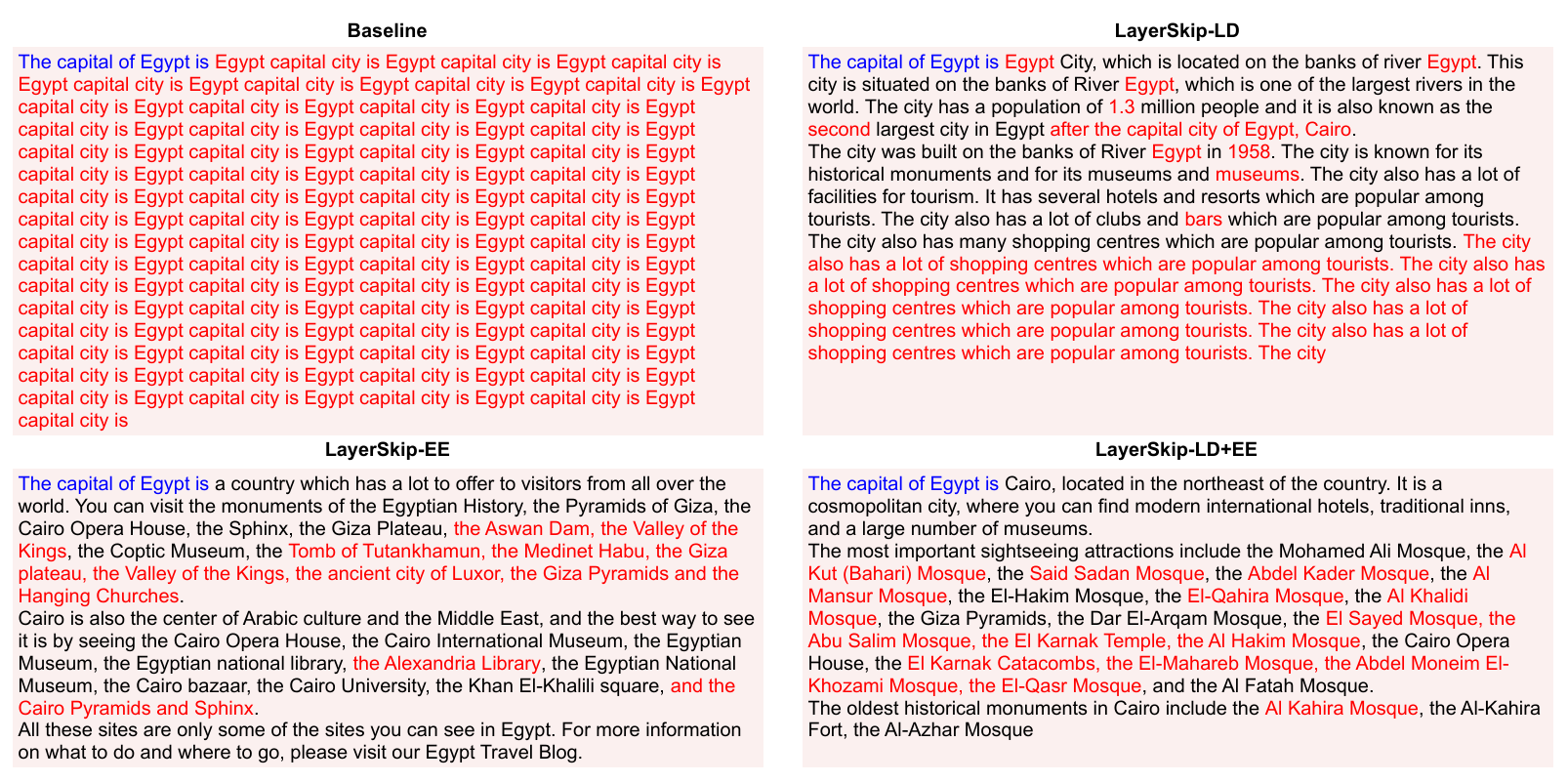}
         \caption{Llama2 7B (that has 32 layers) pretrained from scratch on 26B tokens only, exiting at layer 24.}
         \label{fig:early_exit:generation_pretraining:llama2_7B}
     \end{subfigure}
     \caption{Early exit text generation examples for models pretrained from scratch on 26B tokens with and without \System. \textcolor{blue}{Blue}: The prompt fed into the model. \textcolor{red}{Red}: incorrect phrases or words generated by the model (whether factually or grammatically wrong, or hallucinations). With self-speculative decoding, we fix those incorrect phrases by verifying with remaining layers.}
     \label{fig:early_exit:generation_pretraining}
\end{figure*}

\textbf{Pretraining from Scratch} In Figure~\ref{fig:early_exit:pretraining}, we present our results for Llama2 1.5B and 7B pretrained from scratch on 26B tokens using \System\ on a diverse set of evaluation tasks (see \S~\ref{sec:appendix:experiment_details:evaluation_tasks} for task details) and compare with the same models pretrained on the same number of tokens from scratch without \System. In Figure~\ref{fig:early_exit:generation_pretraining} we show sample text generations for exiting at earlier layers. The results show that introducing our proposed layer dropout configuration and/or early exit loss leads to higher accuracy than the baseline on earlier layers. On the last layer, in some downstream tasks, we do see a slight drop in accuracy compared to baseline, while in other tasks we see our layer dropout and/or early exit configuration leading to higher accuracy. Note that since the models were pretrained on relatively small number of tokens, the evaluation results on some classification tasks were close to random guesses, whether we used \System or not, and hence we have removed them from the results.


\begin{figure*}[!ht]
     \centering
     \begin{subfigure}[b]{0.48\textwidth}
        \centering
        \includegraphics[width=\linewidth]{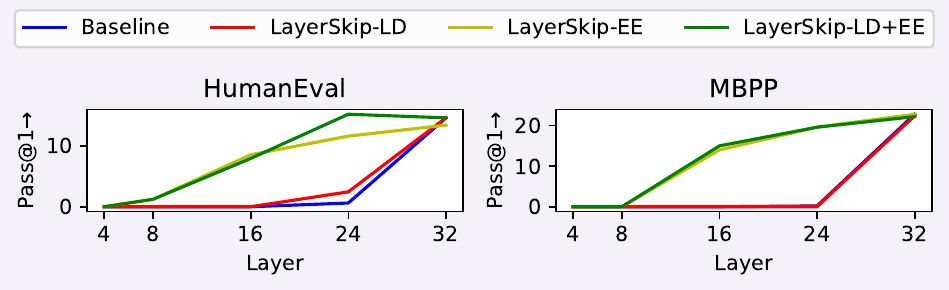}
        \caption{Finetuning Llama1 7B on code.}
        \label{fig:early_exit:finetuning:llama1_7B_code}
     \end{subfigure}
     \hfill
     \begin{subfigure}[b]{0.48\textwidth}
        \centering
        \includegraphics[width=\linewidth]{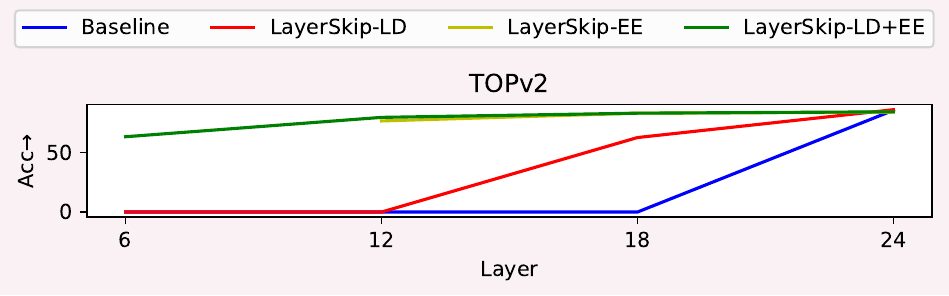}
        \caption{Finetuning Llama1 1.5B on TOPv2 training set.}
        \label{fig:early_exit:finetuning:llama2_1.5B_instruction}
     \end{subfigure}
     \caption{Early exit evaluation of finetuning on domain-specific or task-specific data.}
     \label{fig:early_exit:finetuning}
\end{figure*}

\textbf{Finetuning on Code Data} In Figure~\ref{fig:early_exit:finetuning:llama1_7B_code}, we present our results on 2 coding tasks and compare accuracy to Llama1 7B finetuned on the same number of code tokens without \System. For earlier layers, \System\ is clearly better than the baseline, with layer dropout combined with early exit loss showing a big improvement on one of the 2 tasks. For last layer accuracy, \System\ with both layer dropout and early exit loss has almost the same accuracy as baseline. Note that since this experiment finetuned on specific domain data, we were able to increase $e_{scale}$ to 1.0 (as opposed to $e_{scale}=0.1$ or $0.2$ in the previous two configurations).

\textbf{Finetuning on Task-Specific Dataset} In Figure~\ref{fig:early_exit:finetuning:llama2_1.5B_instruction}, we compare results of fine-tuning our Llama 1.5B model on TOPv2 training set with and without \System. In semantic parsing, correctness requires an exact match (EM) between generated sequence and annotated parse. We find that when removing layers from the baseline model, the model is not able to generate complete and accurate parses resulting in 0 EM. However, with \System, early exit inference improves to 77\% at layer 12. We notice a regression in the final layer reducing accuracy by 3\%. Again, as this configuration finetuned data on a specific task, we were able to set $e_{scale}=1.0$.

\subsection{Self-Speculative Decoding Results}
\label{sec:results:self_speculative_decoding}

We evaluate the self-speculative decoding algorithm introduced in \S \ref{sec:proposed_solution:self_speculative_decoding} on the different trained models. We report quality metrics, EM (exact match) and ROUGE-2~\citet{rouge2} implemented in torchmetrics~\citet{torchmetrics}, token acceptance rate for the self speculation algorithm (how often verification accepts each of the draft tokens), throughput measured as tokens per second averaged over the sampled dataset, and finally a speed up compared to autoregressive decoding. For our early exit and our self-speculative decoding experiments, we specify the layer we exit at in the $E$ column in each table below. We compare with Draft \& Verify~\citet{zhang2023draft} on common models and tasks evaluated in both papers. All experiments were performed with greedy decoding and generated a maximum of 512 tokens for each sample. Following ~\citet{zhang2023draft}, speedup is calculated as the acceleration of average inference time per token compared to the ``Autoregressive'' baseline on the same setting. ``Autoregressive'' experiments use baseline models that were pretrained or finetuned without \System, while ``Early Exit'' and ``Self Speculative'' experiments use our models trained or finetuned with \System. Our implementation leverages HuggingFace~\citet{wolf-etal-2020-transformers}.

\textbf{Continual Pretraining}  In Table~\ref{tab:self_speculative:llama2_continual_pretraining}, we evaluate the continual pre-training of Llama2 7B and 13B with and without \System\ on various tasks: CNN/DM~\citet{nallapati2016abstractive}, XSUM~\citet{xsum} abstractive summarization tasks, and HumanEval~\citet{HumanEval} coding task. The experiments were performed on NVIDIA H100 GPUs. The number of speculations, i.e., the  number of tokens generated in the draft stage in self-speculation, is specified in the $d$ column. We follow \citet{zhang2023draft} and perform CNN/DM with 1-shot abstractive summarization, and XSUM with 0-shot. We obtain speedups between 1.34$\times$ and 2.16$\times$ depending on the model or the task. In general, we observe higher speedups for the smaller Llama2 7B compared to the larger Llama2 13B model. When comparing with Draft \& Verify~\citet{zhang2023draft} on the common model and tasks of both approaches, we are significantly faster on CNN/DM (1.81$\times$ vs. 1.5$\times$) and slightly slower on XSUM (1.34$\times$ vs. 1.48$\times$).
 

\textbf{Pretraining from Scratch} Our results are presented in Table~\ref{tab:self_speculative:llama2_scratch_pretraining}. The experiments were performed on NVIDIA H100 GPUs. We found that an opposite trend to continual pretraining: the bigger model has a bigger speedup than the smaller model, reaching 2.16$\times$ speedup, which exceeds traditional speculative decoding. 

\textbf{Finetuning on Code Data} In Table~\ref{tab:self_speculative:llama1_7B_code_finetuning}, we evaluate our code-finetuned Llama1 7B on HumanEval using 12 speculations, and exit at layer 6 for self speculation \& early exit. The experiments were performed on NVIDIA A100 GPUs. We show speedup of upto 1.82$\times$ with no accuracy drop.

\textbf{Finetuning on Task-Specific Dataset} In Table~\ref{tab:result-ss-topv2} we show results for Llama 1.5B finetuned on TOPv2's training dataset and evaluated on TOPv2 test set. The experiments were performed on NVIDIA H100 GPUs. We present the EM (exact match) on the fully TOPv2 test set, further we sample 1000 samples for latency experiments where we leverage 8 speculations, and generate the next 80 tokens with greedy decoding. With self-speculation, the model was able to achieve high token acceptance rate, ($E=6$: 76.0\%, $E=12$: 97.2\%, $E=18$: 98.9\%) reaching 2.0$\times$ speedup.

\begin{table}[!ht]
\small
\centering
\setlength{\tabcolsep}{4pt}
\begin{NiceTabular}{lcccccc|cccccc}
    \CodeBefore
    \rectanglecolor{greenlight}{1-2}{2-7}
    \rectanglecolor{turquoiselight}{1-8}{2-13}
    \rectanglecolor{white}{3-1}{3-13}
    \rectanglecolor{greenlight}{5-2}{8-7}
    \rectanglecolor{turquoiselight}{5-8}{8-13}
    \rectanglecolor{white}{9-1}{10-13}
    \rectanglecolor{greenlight}{9-2}{14-7}
    \rectanglecolor{turquoiselight}{9-8}{14-13}
    \rectanglecolor{white}{15-1}{16-13}
    \rectanglecolor{greenlight}{17-2}{19-7}
    \rectanglecolor{turquoiselight}{17-8}{19-13}
    \Body
    \toprule
        & \multicolumn{6}{c}{\textbf{Llama2 7B}} & \multicolumn{6}{c}{\textbf{Llama2 13B}} \\
    \toprule
    Generation & \textit{E} & \textit{d} & ROUGE-2 & \shortstack[c]{Token\\Acc.}  & \shortstack[c]{Tokens\\per Sec.} & Speedup & \textit{E} & \textit{d} & ROUGE-2 & \shortstack[c]{Token\\Acc.}  & \shortstack[c]{Tokens\\per Sec.} & Speedup \\
    \midrule
    \multicolumn{6}{l}{\textbf{CNN-DM}} \\
    \multicolumn{6}{l}{One-Shot Abstractive Summarization} \\
    \midrule
    Autoregressive      & -   & -   &   0.079  &   -       &   62.7    &   1.00$\times$             & -   & -   &   0.098   &   -       &   37.2    &   1.00$\times$            \\   
    Early Exit          & 8   & -   &   0.012  &   -       &   232.4   &   -                        & 15  & -   &   0.016   &   -       &   105.5   &   -                       \\      
    Self Speculative    & 8   & 12  &   0.078  &   68.9\%  &   127.9   &   \textbf{1.86$\times$}    & 15  & 12  &   0.098   &   74.5\%  &   70.2    &   \textbf{1.81$\times$}   \\
    \midrule
    Draft and Verify    & n/a & n/a &   n/a    &    n/a    &   n/a     &   n/a                      & -   & -   &  0.107    &   n/a     &   n/a     &   1.56$\times$             \\
    \midrule
    \multicolumn{6}{l}{\textbf{XSUM}} \\
    \multicolumn{6}{l}{Abstractive Summarization} \\
    \midrule
    Autoregressive      & -   & -    &   0.073 &   -       &   63.4    &   1.00$\times$            & -   & -   &  0.124    &   -       &   43.8    &   1.00$\times$              \\ 
    Early Exit          & 8   & -    &   0.002 &   -       &   228.0   &   -                       & 15  & -   &   0.009   &   -       &   110.6   &   -                        \\
    Self Speculative    & 8   & 12   &  0.073  &   54.6\%  &   104.7   &   \textbf{1.54$\times$}   & 15  & 4   &   0.124   &   67.7\%  &   60.5    &   1.34$\times$             \\
    \midrule
    Draft and Verify    & n/a & n/a  &   n/a   &   n/a     &   n/a     &   n/a                     & -   & -   &   0.126   &   n/a     &   n/a     &   \textbf{1.48$\times$}    \\
    \midrule
    \multicolumn{6}{l}{\textbf{HumanEval}} \\
    \multicolumn{6}{l}{Coding} \\
    \midrule
    Autoregressive      & -   & -    &   0.041 &   -       &   62.9    &   1.00$\times$            & -   & -   &   0.055   &   -       &   48.9    &   1.00$\times$ \\
    Early Exit          & 8   & -    &   0.003 &   -       &   225.4   &   -                       & 15  & -   &   0.0005  &   -       &   244.3   &   - \\
    Self Speculative    & 8   & 6    &  0.042  &   67.1\%  &   122.8   &   \textbf{1.83$\times$}   & 7   & 4   &   0.055   &   57.0\%  &   84.2    &   \textbf{1.66$\times$} \\ 
    \bottomrule
\end{NiceTabular}
\caption{\small Generation results for Llama2 continually pretrained with and without LayerSkip.}
\label{tab:self_speculative:llama2_continual_pretraining}
\end{table}

\begin{table}[!ht]
\small
\centering
\setlength{\tabcolsep}{4pt}
\begin{NiceTabular}{lccccc|ccccccc}
    \CodeBefore
    \rectanglecolor{orangelight}{1-2}{2-6}
    \rectanglecolor{redlight}{1-7}{2-11}
    \rectanglecolor{white}{3-1}{3-11}
    \rectanglecolor{orangelight}{5-2}{7-6}
    \rectanglecolor{redlight}{5-7}{7-11}
    \Body
    \toprule
        & \multicolumn{5}{c}{\textbf{Llama2 1.5B - 26B Tokens}} & \multicolumn{5}{c}{\textbf{Llama2 7B - 26B Tokens}} \\
    \toprule
    Generation & \shortstack[l]{\textit{E}} & ROUGE-2 & \shortstack[c]{Token\\Acc.}  & \shortstack[c]{Tokens\\per Sec.} & Speedup & \shortstack[c]{\textit{E}} & ROUGE-2 & \shortstack[c]{Token\\Acc.}  & \shortstack[c]{Tokens\\per Sec.} & Speedup \\
    \midrule
    \multicolumn{6}{l}{\textbf{CNN-DM}} \\
    \multicolumn{6}{l}{One-Shot Abstractive Summarization} \\
    \midrule
    Autoregressive      &   -   &   0.063  &   -       &   91.6    &   1.00$\times$             &   -   &   0.060       &   -       &   64.5       &   1.00$\times$            \\   
    Self Speculative    &   8   &   0.063  &   77.4\%  &   167.4   &   \textbf{1.76$\times$}    &   8   &   0.067       &   77.8\%     &   145.6       &   \textbf{2.16$\times$}   \\
    \bottomrule
\end{NiceTabular}
\caption{\small Generation results for Llama2 pretrained from scratch on 26B tokens with and without LayerSkip.}
\label{tab:self_speculative:llama2_scratch_pretraining}
\end{table}

\begin{table}[!ht]
\parbox{.45\linewidth}{
\small
\centering
\setlength{\tabcolsep}{4pt}
\newcolumntype{p}{>{\columncolor{purplelight}}c}
\begin{tabular}{lpppppp}
    \toprule
    Generation & \textit{E} & ROUGE-2 & \shortstack[c]{Token\\Acc.} & \shortstack[c]{Tokens\\per Sec.} & Speedup \\
    \hline
    Autoregressive & - & 0.0513 &  - & 34 & 1.0$\times$ \\
    Early Exit & 6 & 0.0035 & - & 170 & - \\
    Self Speculative & 6 & 0.0513 & 45\% & 62 & \textbf{1.82$\times$} \\
    \bottomrule
\end{tabular}
\caption{\small Generation results on HumanEval for Llama 7B finetuned on code}
\label{tab:self_speculative:llama1_7B_code_finetuning}
}
\hfill
\parbox{.45\linewidth}{
\small
\centering
\setlength{\tabcolsep}{4pt}
\newcolumntype{i}{>{\columncolor{pinklight}}c}
\begin{tabular}{liiiiii}
    \toprule
    Generation & \textit{E} & EM & \shortstack[c]{Token\\Acc.} & \shortstack[c]{Time per\\Token (ms)} & Speedup \\
    \midrule
    Autoregressive & - & 85.9\% & - & 36 & 1.00$\times$ \\
    \midrule
    Early Exit & 18 & 83.3\% & - & 28 & - \\
    Early Exit & 12 & 79.4\% & - & 19 & - \\
    Early Exit & 6 & 62.9\% & - & 10 & - \\
    \midrule
    Self Speculative & 18 & 82.9\% & 98.9\% & 29 & 1.24$\times$  \\
    Self Speculative & 12 & 82.9\% & 97.6\% & 22 & 1.64$\times$ \\
    Self Speculative & 6 & 82.9\% & 76.0\% & 18 & \textbf{2.0$\times$} \\
    \bottomrule
\end{tabular}
\caption{\small Generation results on TOPv2 task for Llama 1.5B finetuned on TOPv2 training data.}
\label{tab:result-ss-topv2}
}
\end{table}

\section{Ablation Studies}
\label{sec:ablation}

\begin{figure}
    \centering
    \includegraphics[width=0.48\textwidth]{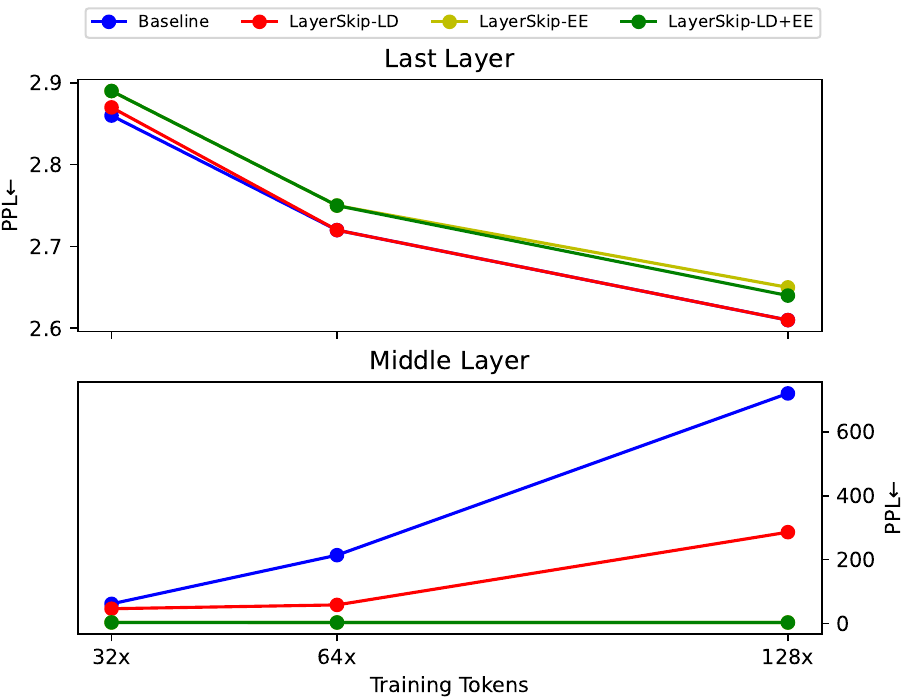}
    \caption{Perplexity on The Stack~\citet{Kocetkov2022TheStack} test set when pretraining Llama 1.5B from scratch with different number of tokens. }
    \label{fig:ablation:scaling_pretraining_tokens}
\end{figure}

\textbf{Scaling with Pretraining Tokens}
\label{sec:ablation:scaling_pretraining_tokens}
In order to understand how the accuracy of last and middle layers change across time when pretraining from scratch, we ran 3 training experiments with different number of tokens on Llama 1.5B and show the results in Figure~\ref{fig:ablation:scaling_pretraining_tokens}. Each experiment trained for 50,000 steps, per device batch size of 4, context window of 4096, but changed the number of GPUs to 32, 64, 128. We plotted the perplexity of a held out split of The Stack dataset on the last layer (layer 24) and the middle layer (layer 12). As expected, perplexity on last layer decreases as we train on more tokens. However, surprisingly, we discover that perplexity on middle layer increases drastically by default in training, unless we apply early exit loss. Layer dropout reduces the increase as well. This could open the door to more research on the dynamics of transformers and the evolution of embeddings in earlier layers to understand why embeddings across layers are close to each other early on in training but diverge drastically as training progresses. This could also present a motivation for our training recipe that has minimal drop in last layer accuracy while significantly improves accuracy of earlier layers.

\textbf{KV Cache in Self-Speculation}
\label{sec:ablation:kv_cache_speculation}
In \S \ref{sec:proposed_solution:self_speculative_decoding:resuing_cache} we introduced the re-use of KV cache as a method for improving model generation speed. We measure its effect in Table~\ref{tab:ablation-kv-cache}. We follow the same inference setup as described in \S \ref{sec:results:self_speculative_decoding}. We find that the use of KV cache is able to consistently save us 9-20 ms per token depending on the task.

\begin{table}[t]
\small
\centering
\setlength{\tabcolsep}{4pt}
\begin{tabular}{lrr}
    \toprule
    Generation & TOPv2 ms/t & CNN/DM ms/t \\
    \midrule
    Self Speculation($E=18$) & \textbf{134} & \textbf{166} \\
    \quad w.o KVQ Reuse &  143  & 182  \\
    Self Speculation($E=12$) & \textbf{104} & \textbf{165} \\
    \quad w.o KVQ Reuse &  110  & 185  \\
    \bottomrule
\end{tabular}
\caption{\small Ablation on re-use of the KV cache and exit query cache. Results are presented on CPU inference.}
\label{tab:ablation-kv-cache}
\end{table}


\section{Limitations}
\label{sec:limitation}
\begin{itemize}[noitemsep, nolistsep, topsep=0pt]
    \item Our self-speculative decoding solution requires finetuning a model or pretraining it with our recipe, while the self-speculative decoding approach propoposed in ~\citet{zhang2023draft} does not require changing a model's weights.
    \item The introduced hyperparameters, $p_{max}$ for layer dropout, $e_{scale}$ and $R$ for early exit, requires tuning in order to avoid a drop in last layer accuracy.
    \item When pretraining with layer dropout from scratch, increasing the learning rate is required to maintain accuracy, and tuning learning rate to get optimal accuracy could be tricky and time consuming.
\end{itemize}

\section{Conclusion}
\label{sec:conclusion}




We have shown that combining layer dropout, early exit loss with curriculum, improves accuracy of early exit during inference. We then developed a novel self-speculative decoding solution that led upto 1.86$\times$ speedup.

We hope this paper will encourage researchers and engineers to adopt the proposed layer dropout and/or early exit loss in pretraining as well as finetuning recipes. For pretraining from scratch, layer dropout can lead to increase in training speed. While early exit loss may introduce an overhead for pretraining, using a large $R$ can make that minimal. For finetuning, we hope that our proposed techniques may be combined with parameter efficient techniques suchs as LoRA~\citet{hu2021lora}.

In the future, we would like to increase the accuracy of early-exit layers in order to obtain better speedups during self-speculative decoding. We can also explore dynamic conditions to determine a different exit layer for each token (like ~\citet{calm}) and hence improve token acceptance rate of self-speculative decoding.

\section*{Acknowledgements}
We would like to thank Volker Seeker, Artem Korenev, and Ilia Kulikov for logistic support, FAIR's clusters' support team members, especially, Henry Estela, Hongsheng Song, Shubho Sengupta, and Nabib Ahmed, for their help in maintaing our clusters, Helen Klein, Kamila Benzina, and Ty Toledano for legal support, Fabian Gloeckle, Andrey Gromov, Francisco Massa, Daniel Haziza, Aaditya Singh, Karen Hambardzumyan, Nicola Cancedda, for discussions.

\clearpage
\newpage
\bibliographystyle{assets/plainnat}
\bibliography{paper}

\begin{thebibliography}{69}
\providecommand{\natexlab}[1]{#1}
\providecommand{\url}[1]{\texttt{#1}}
\expandafter\ifx\csname urlstyle\endcsname\relax
  \providecommand{\doi}[1]{doi: #1}\else
  \providecommand{\doi}{doi: \begingroup \urlstyle{rm}\Url}\fi

\bibitem[Austin et~al.(2021)Austin, Odena, Nye, Bosma, Michalewski, Dohan, Jiang, Cai, Terry, Le, and Sutton]{MBPP}
Jacob Austin, Augustus Odena, Maxwell Nye, Maarten Bosma, Henryk Michalewski, David Dohan, Ellen Jiang, Carrie Cai, Michael Terry, Quoc Le, and Charles Sutton.
\newblock Program synthesis with large language models, 2021.

\bibitem[Bisk et~al.(2020)Bisk, Zellers, Bras, Gao, and Choi]{PIQA}
Yonatan Bisk, Rowan Zellers, Ronan~Le Bras, Jianfeng Gao, and Yejin Choi.
\newblock Piqa: Reasoning about physical commonsense in natural language.
\newblock In \emph{Thirty-Fourth AAAI Conference on Artificial Intelligence}, 2020.

\bibitem[Brown et~al.(2020)Brown, Mann, Ryder, Subbiah, Kaplan, Dhariwal, Neelakantan, Shyam, Sastry, Askell, Agarwal, Herbert-Voss, Krueger, Henighan, Child, Ramesh, Ziegler, Wu, Winter, Hesse, Chen, Sigler, Litwin, Gray, Chess, Clark, Berner, McCandlish, Radford, Sutskever, and Amodei]{GPT3}
Tom Brown, Benjamin Mann, Nick Ryder, Melanie Subbiah, Jared~D Kaplan, Prafulla Dhariwal, Arvind Neelakantan, Pranav Shyam, Girish Sastry, Amanda Askell, Sandhini Agarwal, Ariel Herbert-Voss, Gretchen Krueger, Tom Henighan, Rewon Child, Aditya Ramesh, Daniel Ziegler, Jeffrey Wu, Clemens Winter, Chris Hesse, Mark Chen, Eric Sigler, Mateusz Litwin, Scott Gray, Benjamin Chess, Jack Clark, Christopher Berner, Sam McCandlish, Alec Radford, Ilya Sutskever, and Dario Amodei.
\newblock Language models are few-shot learners.
\newblock In H.~Larochelle, M.~Ranzato, R.~Hadsell, M.F. Balcan, and H.~Lin, editors, \emph{Advances in Neural Information Processing Systems}, volume~33, pages 1877--1901. Curran Associates, Inc., 2020.
\newblock \url{https://proceedings.neurips.cc/paper_files/paper/2020/file/1457c0d6bfcb4967418bfb8ac142f64a-Paper.pdf}.

\bibitem[Cancedda(2024)]{cancedda2024spectral}
Nicola Cancedda.
\newblock Spectral filters, dark signals, and attention sinks, 2024.

\bibitem[Chen et~al.(2023)Chen, Borgeaud, Irving, Lespiau, Sifre, and Jumper]{Chen2023AcceleratingLLSpeculativeSampling}
Charlie Chen, Sebastian Borgeaud, Geoffrey Irving, Jean-Baptiste Lespiau, L.~Sifre, and John~M. Jumper.
\newblock Accelerating large language model decoding with speculative sampling.
\newblock \emph{ArXiv}, abs/2302.01318, 2023.
\newblock \url{https://api.semanticscholar.org/CorpusID:256503945}.

\bibitem[Chen et~al.(2021)Chen, Tworek, Jun, Yuan, de~Oliveira~Pinto, Kaplan, Edwards, Burda, Joseph, Brockman, Ray, Puri, Krueger, Petrov, Khlaaf, Sastry, Mishkin, Chan, Gray, Ryder, Pavlov, Power, Kaiser, Bavarian, Winter, Tillet, Such, Cummings, Plappert, Chantzis, Barnes, Herbert-Voss, Guss, Nichol, Paino, Tezak, Tang, Babuschkin, Balaji, Jain, Saunders, Hesse, Carr, Leike, Achiam, Misra, Morikawa, Radford, Knight, Brundage, Murati, Mayer, Welinder, McGrew, Amodei, McCandlish, Sutskever, and Zaremba]{HumanEval}
Mark Chen, Jerry Tworek, Heewoo Jun, Qiming Yuan, Henrique~Ponde de~Oliveira~Pinto, Jared Kaplan, Harri Edwards, Yuri Burda, Nicholas Joseph, Greg Brockman, Alex Ray, Raul Puri, Gretchen Krueger, Michael Petrov, Heidy Khlaaf, Girish Sastry, Pamela Mishkin, Brooke Chan, Scott Gray, Nick Ryder, Mikhail Pavlov, Alethea Power, Lukasz Kaiser, Mohammad Bavarian, Clemens Winter, Philippe Tillet, Felipe~Petroski Such, Dave Cummings, Matthias Plappert, Fotios Chantzis, Elizabeth Barnes, Ariel Herbert-Voss, William~Hebgen Guss, Alex Nichol, Alex Paino, Nikolas Tezak, Jie Tang, Igor Babuschkin, Suchir Balaji, Shantanu Jain, William Saunders, Christopher Hesse, Andrew~N. Carr, Jan Leike, Josh Achiam, Vedant Misra, Evan Morikawa, Alec Radford, Matthew Knight, Miles Brundage, Mira Murati, Katie Mayer, Peter Welinder, Bob McGrew, Dario Amodei, Sam McCandlish, Ilya Sutskever, and Wojciech Zaremba.
\newblock Evaluating large language models trained on code, 2021.

\bibitem[Chen et~al.(2020)Chen, Ghoshal, Mehdad, Zettlemoyer, and Gupta]{TOPv2}
Xilun Chen, Asish Ghoshal, Yashar Mehdad, Luke Zettlemoyer, and Sonal Gupta.
\newblock Low-resource domain adaptation for compositional task-oriented semantic parsing.
\newblock In Bonnie Webber, Trevor Cohn, Yulan He, and Yang Liu, editors, \emph{Proceedings of the 2020 Conference on Empirical Methods in Natural Language Processing (EMNLP)}, pages 5090--5100, Online, November 2020. Association for Computational Linguistics.
\newblock \doi{10.18653/v1/2020.emnlp-main.413}.
\newblock \url{https://aclanthology.org/2020.emnlp-main.413}.

\bibitem[Chowdhery et~al.(2022)Chowdhery, Narang, Devlin, Bosma, Mishra, Roberts, Barham, Chung, Sutton, Gehrmann, Schuh, Shi, Tsvyashchenko, Maynez, Rao, Barnes, Tay, Shazeer, Prabhakaran, Reif, Du, Hutchinson, Pope, Bradbury, Austin, Isard, Gur-Ari, Yin, Duke, Levskaya, Ghemawat, Dev, Michalewski, Garcia, Misra, Robinson, Fedus, Zhou, Ippolito, Luan, Lim, Zoph, Spiridonov, Sepassi, Dohan, Agrawal, Omernick, Dai, Pillai, Pellat, Lewkowycz, Moreira, Child, Polozov, Lee, Zhou, Wang, Saeta, Diaz, Firat, Catasta, Wei, Meier-Hellstern, Eck, Dean, Petrov, and Fiedel]{chowdhery2022palm}
Aakanksha Chowdhery, Sharan Narang, Jacob Devlin, Maarten Bosma, Gaurav Mishra, Adam Roberts, Paul Barham, Hyung~Won Chung, Charles Sutton, Sebastian Gehrmann, Parker Schuh, Kensen Shi, Sasha Tsvyashchenko, Joshua Maynez, Abhishek Rao, Parker Barnes, Yi~Tay, Noam Shazeer, Vinodkumar Prabhakaran, Emily Reif, Nan Du, Ben Hutchinson, Reiner Pope, James Bradbury, Jacob Austin, Michael Isard, Guy Gur-Ari, Pengcheng Yin, Toju Duke, Anselm Levskaya, Sanjay Ghemawat, Sunipa Dev, Henryk Michalewski, Xavier Garcia, Vedant Misra, Kevin Robinson, Liam Fedus, Denny Zhou, Daphne Ippolito, David Luan, Hyeontaek Lim, Barret Zoph, Alexander Spiridonov, Ryan Sepassi, David Dohan, Shivani Agrawal, Mark Omernick, Andrew~M. Dai, Thanumalayan~Sankaranarayana Pillai, Marie Pellat, Aitor Lewkowycz, Erica Moreira, Rewon Child, Oleksandr Polozov, Katherine Lee, Zongwei Zhou, Xuezhi Wang, Brennan Saeta, Mark Diaz, Orhan Firat, Michele Catasta, Jason Wei, Kathy Meier-Hellstern, Douglas Eck, Jeff Dean, Slav Petrov, and Noah Fiedel.
\newblock Palm: Scaling language modeling with pathways, 2022.

\bibitem[Clark et~al.(2019)Clark, Lee, Chang, Kwiatkowski, Collins, and Toutanova]{BoolQ}
Christopher Clark, Kenton Lee, Ming-Wei Chang, Tom Kwiatkowski, Michael Collins, and Kristina Toutanova.
\newblock Boolq: Exploring the surprising difficulty of natural yes/no questions.
\newblock In \emph{NAACL}, 2019.

\bibitem[Clark et~al.(2018)Clark, Cowhey, Etzioni, Khot, Sabharwal, Schoenick, and Tafjord]{ARC}
Peter Clark, Isaac Cowhey, Oren Etzioni, Tushar Khot, Ashish Sabharwal, Carissa Schoenick, and Oyvind Tafjord.
\newblock Think you have solved question answering? try arc, the ai2 reasoning challenge.
\newblock \emph{ArXiv}, abs/1803.05457, 2018.
\newblock \url{https://api.semanticscholar.org/CorpusID:3922816}.

\bibitem[Cobbe et~al.(2021)Cobbe, Kosaraju, Bavarian, Chen, Jun, Kaiser, Plappert, Tworek, Hilton, Nakano, Hesse, and Schulman]{gsm8k}
Karl Cobbe, Vineet Kosaraju, Mohammad Bavarian, Mark Chen, Heewoo Jun, Lukasz Kaiser, Matthias Plappert, Jerry Tworek, Jacob Hilton, Reiichiro Nakano, Christopher Hesse, and John Schulman.
\newblock Training verifiers to solve math word problems, 2021.

\bibitem[Corro et~al.(2023)Corro, Giorno, Agarwal, Yu, Awadallah, and Mukherjee]{delcorro2023skipdecode}
Luciano~Del Corro, Allie~Del Giorno, Sahaj Agarwal, Bin Yu, Ahmed Awadallah, and Subhabrata Mukherjee.
\newblock Skipdecode: Autoregressive skip decoding with batching and caching for efficient llm inference, 2023.

\bibitem[Deng et~al.(2009)Deng, Dong, Socher, Li, Li, and Fei-Fei]{DenDon09Imagenet}
Jia Deng, Wei Dong, Richard Socher, Li-Jia Li, Kai Li, and Li~Fei-Fei.
\newblock Imagenet: A large-scale hierarchical image database.
\newblock In \emph{Computer Vision and Pattern Recognition, 2009. CVPR 2009. IEEE Conference on}, pages 248--255. IEEE, 2009.
\newblock \url{https://ieeexplore.ieee.org/abstract/document/5206848/}.

\bibitem[Din et~al.(2023)Din, Karidi, Choshen, and Geva]{jumpingToConclusion}
Alexander~Yom Din, Taelin Karidi, Leshem Choshen, and Mor Geva.
\newblock Jump to conclusions: Short-cutting transformers with linear transformations, 2023.

\bibitem[Elbayad et~al.(2020)Elbayad, Gu, Grave, and Auli]{DepthAdaptive}
Maha Elbayad, Jiatao Gu, Edouard Grave, and Michael Auli.
\newblock Depth-adaptive transformer.
\newblock In \emph{In Proc. of ICLR}, 2020.

\bibitem[Fan et~al.(2020)Fan, Grave, and Joulin]{LayerDrop}
Angela Fan, Edouard Grave, and Armand Joulin.
\newblock Reducing transformer depth on demand with structured dropout.
\newblock In \emph{International Conference on Learning Representations}, 2020.
\newblock \url{https://openreview.net/forum?id=SylO2yStDr}.

\bibitem[Ganesan(2018)]{rouge2}
Kavita Ganesan.
\newblock Rouge 2.0: Updated and improved measures for evaluation of summarization tasks, 2018.

\bibitem[Gao et~al.(2020)Gao, Biderman, Black, Golding, Hoppe, Foster, Phang, He, Thite, Nabeshima, Presser, and Leahy]{pile}
Leo Gao, Stella Biderman, Sid Black, Laurence Golding, Travis Hoppe, Charles Foster, Jason Phang, Horace He, Anish Thite, Noa Nabeshima, Shawn Presser, and Connor Leahy.
\newblock The {P}ile: An 800gb dataset of diverse text for language modeling.
\newblock \emph{arXiv preprint arXiv:2101.00027}, 2020.

\bibitem[Geva et~al.(2022)Geva, Caciularu, Wang, and Goldberg]{prediction_saturation}
Mor Geva, Avi Caciularu, Kevin Wang, and Yoav Goldberg.
\newblock Transformer feed-forward layers build predictions by promoting concepts in the vocabulary space.
\newblock In Yoav Goldberg, Zornitsa Kozareva, and Yue Zhang, editors, \emph{Proceedings of the 2022 Conference on Empirical Methods in Natural Language Processing}, pages 30--45, Abu Dhabi, United Arab Emirates, December 2022. Association for Computational Linguistics.
\newblock \doi{10.18653/v1/2022.emnlp-main.3}.
\newblock \url{https://aclanthology.org/2022.emnlp-main.3}.

\bibitem[Goodfellow et~al.(2016)Goodfellow, Bengio, and Courville]{Goodfellow-et-al-2016}
Ian Goodfellow, Yoshua Bengio, and Aaron Courville.
\newblock \emph{Deep Learning}.
\newblock MIT Press, 2016.
\newblock \url{http://www.deeplearningbook.org}.

\bibitem[Hendrycks et~al.(2021{\natexlab{a}})Hendrycks, Burns, Basart, Zou, Mazeika, Song, and Steinhardt]{MMLU}
Dan Hendrycks, Collin Burns, Steven Basart, Andy Zou, Mantas Mazeika, Dawn Song, and Jacob Steinhardt.
\newblock Measuring massive multitask language understanding.
\newblock In \emph{International Conference on Learning Representations}, 2021{\natexlab{a}}.
\newblock \url{https://openreview.net/forum?id=d7KBjmI3GmQ}.

\bibitem[Hendrycks et~al.(2021{\natexlab{b}})Hendrycks, Burns, Kadavath, Arora, Basart, Tang, Song, and Steinhardt]{hendrycksmath2021}
Dan Hendrycks, Collin Burns, Saurav Kadavath, Akul Arora, Steven Basart, Eric Tang, Dawn Song, and Jacob Steinhardt.
\newblock Measuring mathematical problem solving with the math dataset.
\newblock \emph{NeurIPS}, 2021{\natexlab{b}}.

\bibitem[Hooper et~al.(2024)Hooper, Kim, Mohammadzadeh, Genc, Keutzer, Gholami, and Shao]{hooper2024speed}
Coleman Hooper, Sehoon Kim, Hiva Mohammadzadeh, Hasan Genc, Kurt Keutzer, Amir Gholami, and Sophia Shao.
\newblock Speed: Speculative pipelined execution for efficient decoding, 2024.

\bibitem[Hu et~al.(2021)Hu, Shen, Wallis, Allen-Zhu, Li, Wang, Wang, and Chen]{hu2021lora}
Edward~J. Hu, Yelong Shen, Phillip Wallis, Zeyuan Allen-Zhu, Yuanzhi Li, Shean Wang, Lu~Wang, and Weizhu Chen.
\newblock Lora: Low-rank adaptation of large language models, 2021.

\bibitem[Huang et~al.(2016)Huang, Sun, Liu, Sedra, and Weinberger]{stochastic_depth}
Gao Huang, Yu~Sun, Zhuang Liu, Daniel Sedra, and Kilian Weinberger.
\newblock Deep networks with stochastic depth, 2016.

\bibitem[Ioffe and Szegedy(2015)]{batchnorm}
Sergey Ioffe and Christian Szegedy.
\newblock Batch normalization: Accelerating deep network training by reducing internal covariate shift, 2015.

\bibitem[Kembhavi et~al.(2017)Kembhavi, Seo, Schwenk, Choi, Farhadi, and Hajishirzi]{TQA}
Aniruddha Kembhavi, Minjoon Seo, Dustin Schwenk, Jonghyun Choi, Ali Farhadi, and Hannaneh Hajishirzi.
\newblock Are you smarter than a sixth grader? textbook question answering for multimodal machine comprehension.
\newblock \emph{2017 IEEE Conference on Computer Vision and Pattern Recognition (CVPR)}, pages 5376--5384, 2017.
\newblock \url{https://api.semanticscholar.org/CorpusID:1310550}.

\bibitem[Kocetkov et~al.(2022)Kocetkov, Li, Ben~Allal, Li, Mou, Muñoz~Ferrandis, Jernite, Mitchell, Hughes, Wolf, Bahdanau, von Werra, and de~Vries]{Kocetkov2022TheStack}
Denis Kocetkov, Raymond Li, Loubna Ben~Allal, Jia Li, Chenghao Mou, Carlos Muñoz~Ferrandis, Yacine Jernite, Margaret Mitchell, Sean Hughes, Thomas Wolf, Dzmitry Bahdanau, Leandro von Werra, and Harm de~Vries.
\newblock The stack: 3 tb of permissively licensed source code.
\newblock \emph{Preprint}, 2022.

\bibitem[Kwiatkowski et~al.(2019)Kwiatkowski, Palomaki, Redfield, Collins, Parikh, Alberti, Epstein, Polosukhin, Kelcey, Devlin, Lee, Toutanova, Jones, Chang, Dai, Uszkoreit, Le, and Petrov]{NQ}
Tom Kwiatkowski, Jennimaria Palomaki, Olivia Redfield, Michael Collins, Ankur Parikh, Chris Alberti, Danielle Epstein, Illia Polosukhin, Matthew Kelcey, Jacob Devlin, Kenton Lee, Kristina~N. Toutanova, Llion Jones, Ming-Wei Chang, Andrew Dai, Jakob Uszkoreit, Quoc Le, and Slav Petrov.
\newblock Natural questions: a benchmark for question answering research.
\newblock \emph{Transactions of the Association of Computational Linguistics}, 2019.

\bibitem[Lai et~al.(2017)Lai, Xie, Liu, Yang, and Hovy]{lai-etal-2017-race}
Guokun Lai, Qizhe Xie, Hanxiao Liu, Yiming Yang, and Eduard Hovy.
\newblock {RACE}: Large-scale {R}e{A}ding comprehension dataset from examinations.
\newblock In Martha Palmer, Rebecca Hwa, and Sebastian Riedel, editors, \emph{Proceedings of the 2017 Conference on Empirical Methods in Natural Language Processing}, pages 785--794, Copenhagen, Denmark, September 2017. Association for Computational Linguistics.
\newblock \doi{10.18653/v1/D17-1082}.
\newblock \url{https://aclanthology.org/D17-1082}.

\bibitem[Leviathan et~al.(2023)Leviathan, Kalman, and Matias]{yaniv_speculative_decoding}
Yaniv Leviathan, Matan Kalman, and Yossi Matias.
\newblock Fast inference from transformers via speculative decoding.
\newblock In \emph{Proceedings of the 40th International Conference on Machine Learning}, ICML'23. JMLR.org, 2023.

\bibitem[Liu et~al.(2024)Liu, Zhao, Iandola, Lai, Tian, Fedorov, Xiong, Chang, Shi, Krishnamoorthi, Lai, and Chandra]{liu2024mobilellm}
Zechun Liu, Changsheng Zhao, Forrest Iandola, Chen Lai, Yuandong Tian, Igor Fedorov, Yunyang Xiong, Ernie Chang, Yangyang Shi, Raghuraman Krishnamoorthi, Liangzhen Lai, and Vikas Chandra.
\newblock Mobilellm: Optimizing sub-billion parameter language models for on-device use cases, 2024.

\bibitem[Liu et~al.(2022)Liu, Mao, Wu, Feichtenhofer, Darrell, and Xie]{liu2022convnext}
Zhuang Liu, Hanzi Mao, Chao-Yuan Wu, Christoph Feichtenhofer, Trevor Darrell, and Saining Xie.
\newblock A convnet for the 2020s.
\newblock \emph{Proceedings of the IEEE/CVF Conference on Computer Vision and Pattern Recognition (CVPR)}, 2022.

\bibitem[Liu et~al.(2023)Liu, Xu, Jin, Shen, and Darrell]{liu2023dropout_reduces_underfitting}
Zhuang Liu, Zhiqiu Xu, Joseph Jin, Zhiqiang Shen, and Trevor Darrell.
\newblock Dropout reduces underfitting.
\newblock In \emph{International Conference on Machine Learning}, 2023.

\bibitem[Mangrulkar et~al.(2022)Mangrulkar, S, and Sembium]{BE3R}
Sourab Mangrulkar, Ankith~M S, and Vivek Sembium.
\newblock Be3r: Bert-based early-exit using expert routing.
\newblock In \emph{KDD 2022}, 2022.
\newblock \url{https://www.amazon.science/publications/be3r-bert-based-early-exit-using-expert-routing}.

\bibitem[Mihaylov et~al.(2018)Mihaylov, Clark, Khot, and Sabharwal]{OBQA}
Todor Mihaylov, Peter Clark, Tushar Khot, and Ashish Sabharwal.
\newblock Can a suit of armor conduct electricity? a new dataset for open book question answering.
\newblock In \emph{Conference on Empirical Methods in Natural Language Processing}, 2018.
\newblock \url{https://api.semanticscholar.org/CorpusID:52183757}.

\bibitem[Nallapati et~al.(2016)Nallapati, Zhou, dos santos, Gulcehre, and Xiang]{nallapati2016abstractive}
Ramesh Nallapati, Bowen Zhou, Cicero~Nogueira dos santos, Caglar Gulcehre, and Bing Xiang.
\newblock Abstractive text summarization using sequence-to-sequence rnns and beyond, 2016.

\bibitem[Narayan et~al.(2018)Narayan, Cohen, and Lapata]{xsum}
Shashi Narayan, Shay~B. Cohen, and Mirella Lapata.
\newblock Don't give me the details, just the summary! {T}opic-aware convolutional neural networks for extreme summarization.
\newblock In \emph{Proceedings of the 2018 Conference on Empirical Methods in Natural Language Processing}, Brussels, Belgium, 2018.

\bibitem[{Nicki Skafte Detlefsen} et~al.(2022){Nicki Skafte Detlefsen}, {Jiri Borovec}, {Justus Schock}, {Ananya Harsh}, {Teddy Koker}, {Luca Di Liello}, {Daniel Stancl}, {Changsheng Quan}, {Maxim Grechkin}, and {William Falcon}]{torchmetrics}
{Nicki Skafte Detlefsen}, {Jiri Borovec}, {Justus Schock}, {Ananya Harsh}, {Teddy Koker}, {Luca Di Liello}, {Daniel Stancl}, {Changsheng Quan}, {Maxim Grechkin}, and {William Falcon}.
\newblock {TorchMetrics - Measuring Reproducibility in PyTorch}, February 2022.
\newblock \url{https://github.com/Lightning-AI/torchmetrics}.

\bibitem[Nostalgebraist(2020)]{logitlens}
Nostalgebraist.
\newblock "interpreting gpt: the logit lens", August 2020.
\newblock \url{https://www.lesswrong.com/posts/AcKRB8wDpdaN6v6ru/interpreting-gpt-the-logit-lens}.

\bibitem[Panda et~al.(2016)Panda, Sengupta, and Roy]{panda2016conditional}
Priyadarshini Panda, Abhronil Sengupta, and Kaushik Roy.
\newblock Conditional deep learning for energy-efficient and enhanced pattern recognition, 2016.

\bibitem[Phuong and Hutter(2022)]{FormalVerificationTransformer}
Mary Phuong and Marcus Hutter.
\newblock Formal algorithms for transformers.
\newblock \emph{ArXiv}, abs/2207.09238, 2022.
\newblock \url{https://api.semanticscholar.org/CorpusID:250644473}.

\bibitem[Radford et~al.(2019)Radford, Wu, Child, Luan, Amodei, and Sutskever]{gpt2}
Alec Radford, Jeff Wu, Rewon Child, David Luan, Dario Amodei, and Ilya Sutskever.
\newblock Language models are unsupervised multitask learners.
\newblock 2019.

\bibitem[Raposo et~al.(2024)Raposo, Ritter, Richards, Lillicrap, Humphreys, and Santoro]{raposo2024mixtureofdepthsdynamicallyallocatingcompute}
David Raposo, Sam Ritter, Blake Richards, Timothy Lillicrap, Peter~Conway Humphreys, and Adam Santoro.
\newblock Mixture-of-depths: Dynamically allocating compute in transformer-based language models, 2024.
\newblock \url{https://arxiv.org/abs/2404.02258}.

\bibitem[Roemmele et~al.(2011)Roemmele, Bejan, and Gordon]{COPA}
Melissa Roemmele, Cosmin~Adrian Bejan, and Andrew~S. Gordon.
\newblock Choice of plausible alternatives: An evaluation of commonsense causal reasoning.
\newblock In \emph{Logical Formalizations of Commonsense Reasoning, Papers from the 2011 {AAAI} Spring Symposium, Technical Report SS-11-06, Stanford, California, USA, March 21-23, 2011}. {AAAI}, 2011.
\newblock \url{http://www.aaai.org/ocs/index.php/SSS/SSS11/paper/view/2418}.

\bibitem[Rozière et~al.(2023)Rozière, Gehring, Gloeckle, Sootla, Gat, Tan, Adi, Liu, Remez, Rapin, Kozhevnikov, Evtimov, Bitton, Bhatt, Ferrer, Grattafiori, Xiong, Défossez, Copet, Azhar, Touvron, Martin, Usunier, Scialom, and Synnaeve]{codellama}
Baptiste Rozière, Jonas Gehring, Fabian Gloeckle, Sten Sootla, Itai Gat, Xiaoqing~Ellen Tan, Yossi Adi, Jingyu Liu, Tal Remez, Jérémy Rapin, Artyom Kozhevnikov, Ivan Evtimov, Joanna Bitton, Manish Bhatt, Cristian~Canton Ferrer, Aaron Grattafiori, Wenhan Xiong, Alexandre Défossez, Jade Copet, Faisal Azhar, Hugo Touvron, Louis Martin, Nicolas Usunier, Thomas Scialom, and Gabriel Synnaeve.
\newblock Code llama: Open foundation models for code, 2023.

\bibitem[Sakaguchi et~al.(2019)Sakaguchi, Bras, Bhagavatula, and Choi]{winogrande}
Keisuke Sakaguchi, Ronan~Le Bras, Chandra Bhagavatula, and Yejin Choi.
\newblock Winogrande: An adversarial winograd schema challenge at scale, 2019.

\bibitem[{Samsi} et~al.(2023){Samsi}, {Zhao}, {McDonald}, {Li}, {Michaleas}, {Jones}, {Bergeron}, {Kepner}, {Tiwari}, and {Gadepally}]{benchmarkLLM}
Siddharth {Samsi}, Dan {Zhao}, Joseph {McDonald}, Baolin {Li}, Adam {Michaleas}, Michael {Jones}, William {Bergeron}, Jeremy {Kepner}, Devesh {Tiwari}, and Vijay {Gadepally}.
\newblock {From Words to Watts: Benchmarking the Energy Costs of Large Language Model Inference}.
\newblock \emph{arXiv e-prints}, art. arXiv:2310.03003, October 2023.
\newblock \doi{10.48550/arXiv.2310.03003}.

\bibitem[Sap et~al.(2019)Sap, Rashkin, Chen, Le~Bras, and Choi]{SIQA}
Maarten Sap, Hannah Rashkin, Derek Chen, Ronan Le~Bras, and Yejin Choi.
\newblock Social {IQ}a: Commonsense reasoning about social interactions.
\newblock In Kentaro Inui, Jing Jiang, Vincent Ng, and Xiaojun Wan, editors, \emph{Proceedings of the 2019 Conference on Empirical Methods in Natural Language Processing and the 9th International Joint Conference on Natural Language Processing (EMNLP-IJCNLP)}, pages 4463--4473, Hong Kong, China, November 2019. Association for Computational Linguistics.
\newblock \doi{10.18653/v1/D19-1454}.
\newblock \url{https://aclanthology.org/D19-1454}.

\bibitem[Schuster et~al.(2022)Schuster, Fisch, Gupta, Dehghani, Bahri, Tran, Tay, and Metzler]{calm}
Tal Schuster, Adam Fisch, Jai Gupta, Mostafa Dehghani, Dara Bahri, Vinh~Q. Tran, Yi~Tay, and Donald Metzler.
\newblock Confident adaptive language modeling.
\newblock In Alice~H. Oh, Alekh Agarwal, Danielle Belgrave, and Kyunghyun Cho, editors, \emph{Advances in Neural Information Processing Systems}, 2022.
\newblock \url{https://openreview.net/forum?id=uLYc4L3C81A}.

\bibitem[Shim et~al.(2021)Shim, Choi, Sung, and Choi]{headpruning_layerwise}
Kyuhong Shim, Iksoo Choi, Wonyong Sung, and Jungwook Choi.
\newblock Layer-wise pruning of transformer attention heads for efficient language modeling.
\newblock In \emph{2021 18th International SoC Design Conference (ISOCC)}, pages 357--358, 2021.
\newblock \doi{10.1109/ISOCC53507.2021.9613933}.

\bibitem[Srivastava et~al.(2014)Srivastava, Hinton, Krizhevsky, Sutskever, and Salakhutdinov]{GeoffHintonDropout}
Nitish Srivastava, Geoffrey Hinton, Alex Krizhevsky, Ilya Sutskever, and Ruslan Salakhutdinov.
\newblock Dropout: A simple way to prevent neural networks from overfitting.
\newblock \emph{Journal of Machine Learning Research}, 15\penalty0 (56):\penalty0 1929--1958, 2014.
\newblock \url{http://jmlr.org/papers/v15/srivastava14a.html}.

\bibitem[Teerapittayanon et~al.(2017)Teerapittayanon, McDanel, and Kung]{teerapittayanon2017branchynet}
Surat Teerapittayanon, Bradley McDanel, and H.~T. Kung.
\newblock Branchynet: Fast inference via early exiting from deep neural networks, 2017.

\bibitem[Touvron et~al.(2023{\natexlab{a}})Touvron, Lavril, Izacard, Martinet, Lachaux, Lacroix, Rozière, Goyal, Hambro, Azhar, Rodriguez, Joulin, Grave, and Lample]{llama1}
Hugo Touvron, Thibaut Lavril, Gautier Izacard, Xavier Martinet, Marie-Anne Lachaux, Timothée Lacroix, Baptiste Rozière, Naman Goyal, Eric Hambro, Faisal Azhar, Aurelien Rodriguez, Armand Joulin, Edouard Grave, and Guillaume Lample.
\newblock Llama: Open and efficient foundation language models, 2023{\natexlab{a}}.

\bibitem[Touvron et~al.(2023{\natexlab{b}})Touvron, Martin, Stone, Albert, Almahairi, Babaei, Bashlykov, Batra, Bhargava, Bhosale, Bikel, Blecher, Ferrer, Chen, Cucurull, Esiobu, Fernandes, Fu, Fu, Fuller, Gao, Goswami, Goyal, Hartshorn, Hosseini, Hou, Inan, Kardas, Kerkez, Khabsa, Kloumann, Korenev, Koura, Lachaux, Lavril, Lee, Liskovich, Lu, Mao, Martinet, Mihaylov, Mishra, Molybog, Nie, Poulton, Reizenstein, Rungta, Saladi, Schelten, Silva, Smith, Subramanian, Tan, Tang, Taylor, Williams, Kuan, Xu, Yan, Zarov, Zhang, Fan, Kambadur, Narang, Rodriguez, Stojnic, Edunov, and Scialom]{llama2}
Hugo Touvron, Louis Martin, Kevin Stone, Peter Albert, Amjad Almahairi, Yasmine Babaei, Nikolay Bashlykov, Soumya Batra, Prajjwal Bhargava, Shruti Bhosale, Dan Bikel, Lukas Blecher, Cristian~Canton Ferrer, Moya Chen, Guillem Cucurull, David Esiobu, Jude Fernandes, Jeremy Fu, Wenyin Fu, Brian Fuller, Cynthia Gao, Vedanuj Goswami, Naman Goyal, Anthony Hartshorn, Saghar Hosseini, Rui Hou, Hakan Inan, Marcin Kardas, Viktor Kerkez, Madian Khabsa, Isabel Kloumann, Artem Korenev, Punit~Singh Koura, Marie-Anne Lachaux, Thibaut Lavril, Jenya Lee, Diana Liskovich, Yinghai Lu, Yuning Mao, Xavier Martinet, Todor Mihaylov, Pushkar Mishra, Igor Molybog, Yixin Nie, Andrew Poulton, Jeremy Reizenstein, Rashi Rungta, Kalyan Saladi, Alan Schelten, Ruan Silva, Eric~Michael Smith, Ranjan Subramanian, Xiaoqing~Ellen Tan, Binh Tang, Ross Taylor, Adina Williams, Jian~Xiang Kuan, Puxin Xu, Zheng Yan, Iliyan Zarov, Yuchen Zhang, Angela Fan, Melanie Kambadur, Sharan Narang, Aurelien Rodriguez, Robert Stojnic, Sergey Edunov, and Thomas
  Scialom.
\newblock Llama 2: Open foundation and fine-tuned chat models, 2023{\natexlab{b}}.

\bibitem[Vaswani et~al.(2017)Vaswani, Shazeer, Parmar, Uszkoreit, Jones, Gomez, Kaiser, and Polosukhin]{AttentionIsAllYouNeedVaswani}
Ashish Vaswani, Noam Shazeer, Niki Parmar, Jakob Uszkoreit, Llion Jones, Aidan~N Gomez, \L~ukasz Kaiser, and Illia Polosukhin.
\newblock Attention is all you need.
\newblock In I.~Guyon, U.~Von Luxburg, S.~Bengio, H.~Wallach, R.~Fergus, S.~Vishwanathan, and R.~Garnett, editors, \emph{Advances in Neural Information Processing Systems}, volume~30. Curran Associates, Inc., 2017.
\newblock \url{https://proceedings.neurips.cc/paper_files/paper/2017/file/3f5ee243547dee91fbd053c1c4a845aa-Paper.pdf}.

\bibitem[Voita et~al.(2019)Voita, Sennrich, and Titov]{voita2019bottomup}
Elena Voita, Rico Sennrich, and Ivan Titov.
\newblock The bottom-up evolution of representations in the transformer: A study with machine translation and language modeling objectives, 2019.

\bibitem[Voita et~al.(2023)Voita, Ferrando, and Nalmpantis]{voita2023neurons}
Elena Voita, Javier Ferrando, and Christoforos Nalmpantis.
\newblock Neurons in large language models: Dead, n-gram, positional, 2023.

\bibitem[Wolf et~al.(2020)Wolf, Debut, Sanh, Chaumond, Delangue, Moi, Cistac, Rault, Louf, Funtowicz, Davison, Shleifer, von Platen, Ma, Jernite, Plu, Xu, Scao, Gugger, Drame, Lhoest, and Rush]{wolf-etal-2020-transformers}
Thomas Wolf, Lysandre Debut, Victor Sanh, Julien Chaumond, Clement Delangue, Anthony Moi, Pierric Cistac, Tim Rault, Rémi Louf, Morgan Funtowicz, Joe Davison, Sam Shleifer, Patrick von Platen, Clara Ma, Yacine Jernite, Julien Plu, Canwen Xu, Teven~Le Scao, Sylvain Gugger, Mariama Drame, Quentin Lhoest, and Alexander~M. Rush.
\newblock Transformers: State-of-the-art natural language processing.
\newblock In \emph{Proceedings of the 2020 Conference on Empirical Methods in Natural Language Processing: System Demonstrations}, pages 38--45, Online, October 2020. Association for Computational Linguistics.
\newblock \url{https://www.aclweb.org/anthology/2020.emnlp-demos.6}.

\bibitem[Xia et~al.(2023)Xia, Zheng, Li, Zhuang, Zhou, Qiu, Li, Lin, and Song]{xia2023flashllm}
Haojun Xia, Zhen Zheng, Yuchao Li, Donglin Zhuang, Zhongzhu Zhou, Xiafei Qiu, Yong Li, Wei Lin, and Shuaiwen~Leon Song.
\newblock Flash-llm: Enabling cost-effective and highly-efficient large generative model inference with unstructured sparsity, 2023.

\bibitem[Xiao et~al.(2023)Xiao, Lin, Seznec, Wu, Demouth, and Han]{xiao2023smoothquant}
Guangxuan Xiao, Ji~Lin, Mickael Seznec, Hao Wu, Julien Demouth, and Song Han.
\newblock {S}mooth{Q}uant: Accurate and efficient post-training quantization for large language models.
\newblock In \emph{Proceedings of the 40th International Conference on Machine Learning}, 2023.

\bibitem[Xin et~al.(2021)Xin, Tang, Yu, and Lin]{xin-etal-2021-berxit}
Ji~Xin, Raphael Tang, Yaoliang Yu, and Jimmy Lin.
\newblock {BER}xi{T}: Early exiting for {BERT} with better fine-tuning and extension to regression.
\newblock In Paola Merlo, Jorg Tiedemann, and Reut Tsarfaty, editors, \emph{Proceedings of the 16th Conference of the European Chapter of the Association for Computational Linguistics: Main Volume}, pages 91--104, Online, April 2021. Association for Computational Linguistics.
\newblock \doi{10.18653/v1/2021.eacl-main.8}.
\newblock \url{https://aclanthology.org/2021.eacl-main.8}.

\bibitem[Zellers et~al.(2019)Zellers, Holtzman, Bisk, Farhadi, and Choi]{HellaSwag}
Rowan Zellers, Ari Holtzman, Yonatan Bisk, Ali Farhadi, and Yejin Choi.
\newblock {H}ella{S}wag: Can a machine really finish your sentence?
\newblock In Anna Korhonen, David Traum, and Llu{\'\i}s M{\`a}rquez, editors, \emph{Proceedings of the 57th Annual Meeting of the Association for Computational Linguistics}, pages 4791--4800, Florence, Italy, July 2019. Association for Computational Linguistics.
\newblock \doi{10.18653/v1/P19-1472}.
\newblock \url{https://aclanthology.org/P19-1472}.

\bibitem[Zeng et~al.(2023)Zeng, Du, Wang, Xu, Lei, Chen, and Cui]{zeng2023learningskiplanguagemodeling}
Dewen Zeng, Nan Du, Tao Wang, Yuanzhong Xu, Tao Lei, Zhifeng Chen, and Claire Cui.
\newblock Learning to skip for language modeling, 2023.
\newblock \url{https://arxiv.org/abs/2311.15436}.

\bibitem[Zhang et~al.(2023)Zhang, Wang, Li, Shou, Chen, Chen, and Mehrotra]{zhang2023draft}
Jun Zhang, Jue Wang, Huan Li, Lidan Shou, Ke~Chen, Gang Chen, and Sharad Mehrotra.
\newblock Draft \& verify: Lossless large language model acceleration via self-speculative decoding, 2023.

\bibitem[Zhang et~al.(2019)Zhang, Tan, Song, Chen, Bao, and Ma]{SCAN}
Linfeng Zhang, Zhanhong Tan, Jiebo Song, Jingwei Chen, Chenglong Bao, and Kaisheng Ma.
\newblock Scan: A scalable neural networks framework towards compact and efficient models.
\newblock In H.~Wallach, H.~Larochelle, A.~Beygelzimer, F.~d\textquotesingle Alch\'{e}-Buc, E.~Fox, and R.~Garnett, editors, \emph{Advances in Neural Information Processing Systems}, volume~32. Curran Associates, Inc., 2019.
\newblock \url{https://proceedings.neurips.cc/paper_files/paper/2019/file/934b535800b1cba8f96a5d72f72f1611-Paper.pdf}.

\bibitem[Zhang and He(2020)]{PLD}
Minjia Zhang and Yuxiong He.
\newblock Accelerating training of transformer-based language models with progressive layer dropping.
\newblock In H.~Larochelle, M.~Ranzato, R.~Hadsell, M.F. Balcan, and H.~Lin, editors, \emph{Advances in Neural Information Processing Systems}, volume~33, pages 14011--14023. Curran Associates, Inc., 2020.
\newblock \url{https://proceedings.neurips.cc/paper_files/paper/2020/file/a1140a3d0df1c81e24ae954d935e8926-Paper.pdf}.

\bibitem[Zhu et~al.(2023)Zhu, Li, Liu, Ma, and Wang]{zhu2023survey}
Xunyu Zhu, Jian Li, Yong Liu, Can Ma, and Weiping Wang.
\newblock A survey on model compression for large language models, 2023.

\bibitem[Çöplü et~al.(2023)Çöplü, Loedi, Bendiken, Makohin, Bouw, and Cobb]{coplu2023performance}
Tolga Çöplü, Marc Loedi, Arto Bendiken, Mykhailo Makohin, Joshua~J. Bouw, and Stephen Cobb.
\newblock A performance evaluation of a quantized large language model on various smartphones, 2023.

\end{thebibliography}

\clearpage
\newpage
\beginappendix

\section{Appendix}
\label{sec:appendix}

\subsection{Experiment Details}
\label{sec:appendix:experiment_details}
We provide details of training configuration and hyperparameters for each of our experiments in Table~\ref{tab:experiments}.
\begin{table*}[htbp]
\small
\centering
\setlength{\tabcolsep}{4pt}
\begin{tabular}{lccccc}
    \toprule
    Experiment                          & Model         & \shortstack[l]{Batch\\Size}   & Steps         & GPUs \\
    \midrule
    Continual Pretraining               & Llama2 7B     & 4                             & \num{50e3}    & 64 A100 80 GB \\
                                        & Llama2 13B    & 4                             & \num{50e3}    & 64 A100 80 GB \\
    Pretraining from Scratch            & Llama 1.5B    & 4                             & \num{50e3}    & 32 A100 30 GB \\
                                        & Llama2 7B     & 4                             & \num{50e3}    & 32 A100 30 GB \\
    Finetuning on Code Data             & Llama1 7B     & 4                             & \num{10e3}    & 32 A100 80 GB  \\
    Finetuning on Task-Specific Dataset & Llama 1.5B    & 32                            & \num{5.8e3}   & 8 A100 80 GB \\
    \bottomrule
\end{tabular}
\caption{\small Training Hyperparameters and Configuration of Experiments}
\label{tab:experiments}
\end{table*}

When pretraining from scratch, layer dropout leads to higher accuracy when trained on higher learning rate ~\citet{GeoffHintonDropout}. Therefore, we show learning rates of each experiment with and without layer dropout separately in Table~\ref{tab:learning_rates}.

\begin{table*}[htbp]
\small
\centering
\setlength{\tabcolsep}{4pt}
\begin{tabular}{lcccc}
    \toprule
    Experiment                          & Model         & Dropout       & Initial Learning Rate \\
    \midrule
    Continual Pretraining               & Llama2 7B    & \checkmark     & \num{3e-5}    \\  
                                        & Llama2 13B   & \checkmark     & \num{2e-5}    \\  
    Pretraining from Scratch            & Llama 1.5B   &                & \num{4e-4}    \\
                                        & Llama 1.5B   & \checkmark     & \num{8e-4}    \\
                                        & Llama2 7B    &                & \num{3e-4}    \\
                                        & Llama2 7B    & \checkmark     & \num{8e-4}    \\
    Finetuning on Code Data             & Llama1 7B    &                & \num{1e-4}    \\
                                        & Llama1 7B    & \checkmark     & \num{1e-4}    \\
    Finetuning on Task-Specific Dataset & Llama 1.5B   &                & \num{1e-4}    \\
                                        & Llama 1.5B   & \checkmark     & \num{1e-4}    \\
    \bottomrule
\end{tabular}
\caption{\small Learning Rates of Experiments}
\label{tab:learning_rates}
\end{table*}

\subsubsection{Model Architectures}
\label{sec:appendix:experiment_details:architectures}
We provide details of architectures of different models in Table~\ref{tab:architectures}.

\begin{table}[h]
\small
\centering
\setlength{\tabcolsep}{4pt}
\begin{tabular}{lrrrrr}
    \toprule
    Model & Dim & Heads & Layers & Context \\
    \midrule
    Llama 1.5B & 2048 & 16 & 24 & 4096 \\
    \shortstack[l]{Llama1 7B \\ ~\citet{llama1}} & 4096 & 16 & 32 & 2048 \\
    \shortstack[l]{Llama2 7B \\ ~\citet{llama1}} & 4096 & 16 & 32 & 4096 \\
    \shortstack[l]{Llama2 13B \\ ~\citet{llama2}} & 5120 & 40 & 40 & 4096 \\
    \bottomrule
\end{tabular}
\caption{\small Model Architectures}
\label{tab:architectures}
\end{table}

\subsubsection{Evaluation Tasks}
\label{sec:appendix:experiment_details:evaluation_tasks}
We have evaluated our language models on a wide range of tasks. For the sake of discussions in \S ~\ref{sec:results:early_exit}, we categorize the tasks into:
\begin{itemize}
    \item \textbf{``Classification'' Tasks:} where the model responds with one out of pre-defined answers, e.g., multiple-choice questions, or questions whose answers are either ``True'' or ``False'':
    \begin{itemize}
        \item \textbf{Common Sense Reasoning Tasks}
        \begin{itemize}
            \item \textbf{BoolQ}~\citet{BoolQ}
            \item \textbf{PIQA} (Physical Interaction Question Answering)~\citet{PIQA}
            \item \textbf{SIQA} (Social Interaction Question Answering)~\citet{SIQA}
            \item \textbf{HellaSwag}~\citet{HellaSwag}
            \item \textbf{Winogrande 1.1}~\citet{winogrande}
            \item \textbf{ARC} (Abstraction and Reasoning Corpus)~\citet{ARC}
            \begin{itemize}
                \item ARC Challenge
                \item ARC Easy
            \end{itemize}
            \item \textbf{OBQA} (Open Book Question Answers)~\citet{OBQA}
            \item \textbf{COPA} (Choice Of Plausible Alternatives)~\citet{COPA}
        \end{itemize}
        \item \textbf{RACE} (ReAding Comprehension dataset from Examinations)~\citet{lai-etal-2017-race}
        \begin{itemize}
            \item RACE Middle
            \item RACE High
        \end{itemize}
        \item \textbf{MMLU} (Massive Multitask Language Understanding)~\citet{MMLU}
    \end{itemize}
    \item \textbf{``Generation'' Tasks:} where the model responds with an open-ended sequence of tokens and we evaluate either the exact match of the tokens with the reference answer, or, in the case of code, run and test it on a compiler or interpreter.
    \begin{itemize}
        \item \textbf{Question Answering}
        \begin{itemize}
            \item \textbf{NQ} (Natural Questions)~\citet{NQ}
            \item \textbf{TQA} (Textbook Question Answering)~\citet{TQA}
        \end{itemize}
        \item \textbf{Mathematics}
        \begin{itemize}
            \item \textbf{MATH}~\citet{hendrycksmath2021}
            \item \textbf{GSM8K}~\citet{gsm8k}
        \end{itemize}
        \item \textbf{Code Generation}
        \begin{itemize}
            \item \textbf{HumanEval}~\citet{HumanEval}
            \item \textbf{MBPP} (Mostly Basic Python Problems Dataset)~\citet{MBPP}
        \end{itemize}
    \end{itemize}
\end{itemize}

We also evaluate perplexity on held out test sets on the following datasets:
\begin{itemize}
    \item \textbf{The Stack}, a coding dataset~\citet{Kocetkov2022TheStack}
    \item \textbf{Books} ~\citet{pile}
    \item \textbf{Wikipedia}
\end{itemize}


\subsection{Additional Results}
\label{sec:appendix:additional_results}

\subsubsection{Self-Speculative Decoding Results}
\label{sec:appendix:additional_results:self_speculative_decoding}

\textbf{CPU Inference Experiments}
We conduct our task specific fine-tuning on Llama 1.5B to measure decoding performance on CPU as well, showing a near 2$\times$ speed up on CPU as well, presented in Table \ref{tab:appendix-ss-topv2-cpu}. We conduct our experiments using the first 100 samples from the TOPv2 test set, leveraging 7 speculations, generating the next 50 tokens with greedy decoding. 

\begin{table}[htbp]
\small
\centering
\setlength{\tabcolsep}{4pt}
\begin{tabular}{lrrrr}
    \toprule
    Generation & EM & Acceptance & \shortstack[c]{Time per\\Token (ms)} \\
    \midrule
    Autoregressive & 85.39 & - & 165 \\
    \midrule
    \multicolumn{4}{l}{Early Exit} \\
    \quad $E=18$ & 82.0 & - & 124 \\
    \quad $E=12$ & 77.2 & - & 84 \\
    \quad $E=6$ & 29.8 & - & 44 \\
    \midrule
    \multicolumn{4}{l}{Self Speculation} \\
    \quad $E=18$ & 82.9 & 99 & 134 \\
    \quad $E=12$ & 82.9 & 97 & 104 \\
    \quad $E=6$ & 82.9 & 76 & 87 \\
    \bottomrule
\end{tabular}
\caption{\small Generation results on CPU for TOPv2 task for small Llama-like finetuned on TOPv2 training data.}
\label{tab:appendix-ss-topv2-cpu}
\end{table}

\subsection{Ablation Studies}
\label{sec:appendix:ablation_studies}
\begin{figure}[hbtp]
    \centering
    \includegraphics[width=0.75\linewidth]{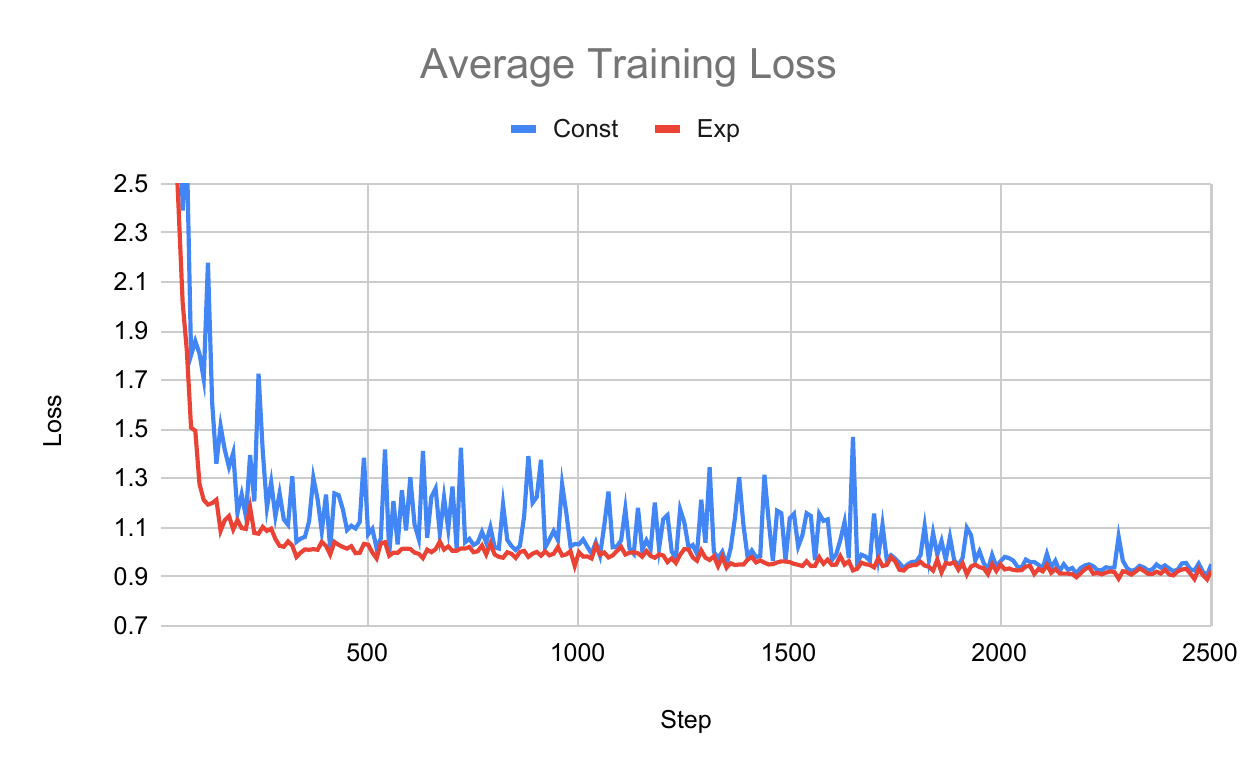}
    \caption{Training loss using different layer dropout configurations. ``Const'' refers to equal dropout on all layers equal to 0.0889, and ``Exp'' refers to dropout exponentially increasing from 0 at the first layer to 0.2 at the last layer. Both configurations have equivalent average dropout across all layers.}
    \label{fig:ablation_studies:layer_drop_config_ablate}
\end{figure}

\textbf{Layer Dropout Configurations}: In Figure~\ref{fig:ablation_studies:layer_drop_config_ablate} we show that our layer dropout configuration leads to lower loss compared to a constant layer dropout across all layers with the same average value.



\newpage
\subsection{Self Speculation Pseudo Code}
\label{sec:appendix:speculation_algo}
Below we share pseudo code for implementing self speculation


\lstset{basicstyle=\tiny\ttfamily,breaklines=true}
\lstset{framextopmargin=50pt,frame=bottomline}
\begin{lstlisting}[language=python]
def self_speculate(
    model,
    input,
    num_speculations,
):
    """
    Input Arguments:
    
    model: Decoder LLM with L layers, supports 2 main
    functions:
    
    * forward_early: computes inference with
    the first E layers of the transformer,
    saves KV states and the exit layer query cache (kvq)
    
     * forward_remainder: computes verification
    with the last L-E layers reusing the kvq cache
    
    input: the input prompt sequence for the model

    num_speculations: the number of speculations from the draft forward pass
    """
    # output contains the generations
    # kvq_cache is the kv cache for generation
    # with the addition of the exit layers query
    # cache to speed up verification
    output, kvq_cache = []
    
    # continue with speculative generation
    # until the maximum sequence length is hit
    while len(output) < max_tokens:
        # produce `num_speculation` draft tokens
        # using the first N layers of the model
        draft_tokens = []
    
        draft_input = input
    
        # conduct num_speculations drafts
        # to produce the draft sequence
        for i in range(num_speculations):
            draft_token = model.forward_e(
                draft_input, kvq_cache
            )
            draft_tokens.append(draft_token)
            draft_input = draft_token
    
        # verify each of the predictions with the
        # full model inference using a single 
        # forward passs
        verified_tokens = model.forward_remainder(
            input + draft_tokens, kvq_cache
        )
    
        # find the number of accepted tokens
        # by comparing the matches between the draft
        # and verified tokens
        matched_tokens = match(
            draft_tokens, verified_tokens)
    
        output.extend(matched_tokens)
        # the last matched token doesn't have a cache
        # and needs to be added
        input = matched_tokens[-1]
    
        # the kvq_cache needs to be updated to remove
        # the last token
        kvq_cache = crop(
            kvq_cache, output_len-1)
    return output
\end{lstlisting}



\end{document}